\documentclass[review, number]{elsarticle}%document class file
\usepackage{amssymb,natbib}

\usepackage{booktabs}
\usepackage{amsmath}
\usepackage{lineno}
\usepackage{times}
\usepackage{latexsym}
\usepackage{url}
\usepackage[utf8]{inputenc}

\usepackage{graphicx} % Required for resizebox
\usepackage{tcolorbox} % For a cleaner box design

\usepackage{microtype}
\usepackage{inconsolata}

\usepackage{graphicx}
\usepackage{afterpage}

\usepackage{amsmath}

\usepackage{xcolor}
\usepackage{enumitem}

\usepackage{multirow}
\usepackage{array}
\usepackage{longtable}
\usepackage{afterpage}
\usepackage{algorithm}
\usepackage{algpseudocode}
\usepackage{amsmath}
\usepackage{array}
\usepackage{graphicx} % Optional, for adjusting box heights
\usepackage{caption}
\usepackage{enumitem}
\usepackage[normalem]{ulem}
\usepackage{tcolorbox}
\usepackage{tabularx}
\usepackage{setspace}
\usepackage{xr}
\usepackage[table]{xcolor}
\journal{npj Artificial Intelligence}

\begin{document}

\begin{frontmatter}



\title{ESGSenticNet: A Neurosymbolic Knowledge Base \\for Corporate Sustainability Analysis}

\author[1,3]{Keane Ong} %% Author name
\author[4]{Rui Mao}
\author[1]{Deeksha Varshney}
\author[2,3]{Frank Xing}
\author[5]{Ranjan Satapathy}
\author[3,6,7]{Johan Sulaeman}
\author[4]{Erik Cambria}
\author[1,3,6,8]{Gianmarco Mengaldo}

% \author[1,3,6,8]{Gianmarco Mengaldo\corref{cor1}}
% \cortext[cor1]{corresponding author: mpegim@nus.edu.sg}

\affiliation[1]{organization={College of Design and Engineering, National University of Singapore},%Department and Organization
            addressline={9 Engineering Drive 1}, 
            postcode={117575}, 
            country={Singapore}}

\affiliation[2]{organization={School of Computing, National University of Singapore},%Department and Organization
            addressline={13 Computing Drive}, 
            postcode={117417}, 
            country={Singapore}}

\affiliation[3]{organization={Asian Institute of Digital Finance, National University of Singapore},%Department and Organization
            addressline={Innovation 4.0, 3 Research Link, \#04-03}, 
            postcode={117602}, 
            country={Singapore}}

\affiliation[4]{organization={College of Computing and Data Science, Nanyang Technological University},%Department and Organization
            addressline={50 Nanyang Ave}, 
            postcode={639798}, 
            country={Singapore}}
            
\affiliation[5]{organization={Institute of High Performance Computing, Agency for Science, Technology
and Research},%Department and Organization
            addressline={Fusionopolis Way, \#16-16 Connexis}, 
            postcode={138632}, 
            country={Singapore}}

\affiliation[6]{organization={Sustainable and Green Finance Institute, National University of Singapore},%Department and Organization
            addressline={Innovation 4.0, 3 Research Link, \#02-02}, 
            postcode={117602}, 
            country={Singapore}}

\affiliation[7]{organization={NUS Business School, National University of Singapore},%Department and Organization
            addressline={Mochtar Riady Building, 15 Kent Ridge Dr, Mocthar Riady Building}, 
            postcode={119245}, 
            country={Singapore}}

\affiliation[8]{organization={Honorary Research Fellow, Imperial College London},
addressline={Exhibition Road, London SW7 2AZ},
country={United Kingdom}%Department and Organization
}
% Abstract
\begin{abstract}
Identifying Environmental, Social, and Governance (ESG) topics in sustainability reports, or ESG topic analysis, is essential for assessing a company's ESG performance. 
While commonly used for this task, existing NLP approaches face three key challenges: \textit{irrelevance}, the extracted topics are unrelated to ESG, \textit{immateriality}, the extracted topics lack meaningful value, and \textit{limited organisation}, the extracted topics have poor alignment with ESG frameworks.

To address these challenges, we introduce ESGSenticNet, a publicly available knowledge base built from 1,679 SGX sustainability reports. ESGSenticNet is constructed using a neurosymbolic framework—combining symbolic reasoning (structured parsing rules) with neural language models (GPT-4o)—to identify and categorise ESG-related concepts. The process begins by extracting concepts (or meaningful phrases) from reports using parsing methods, which are then labelled according to a hierarchical ESG taxonomy using GPT-4o. These labelled concepts are further expanded using a graph-based method that propagates labels to similar concepts without requiring additional manual annotation. The final knowledge base consists of 44,232 structured entries, or “triples,” in the format (concept, relation, ESG topic)—e.g., (“halve carbon emission”, supports, “emissions control”)—which serve as ESG lexicons for topic analysis. This allows ESG topics to be identified by matching concepts found in reports to their associated topics.

We evaluate ESGSenticNet on 319 sustainability reports against existing NLP methods. 
ESGSenticNet outperforms baselines, producing topic terms with 76\% ESG-relatedness (+26\% improvement) and 84\% ESG action-orientation (+34\% improvement), while capturing a high number of unique ESG terms (359). 
Human evaluation of 500 concepts shows that ESGSenticNet achieves 86\%+ accuracy in categorising concepts under correct ESG topics. 
This suggests that during topic analysis, ESGSenticNet shows a strong degree of reliability in aligning topic terms with their intended topics. 
Additionally, ESGSenticNet's deployment requires no training, making it accessible for stakeholders regardless of resources or expertise. Our key contributions include: (1) developing a neurosymbolic framework for deriving sustainability lexicons, which can be potentially extended to other contexts such as climate risk disclosures; (2) introducing new metrics for ESG topic analysis, which we argue are more meaningful for the task at hand; and (3) releasing ESGSenticNet as a public tool that effectively addresses the core challenges of \textit{irrelevance}, \textit{immateriality}, and \textit{limited organisation} in ESG topic analysis.
\end{abstract}


\end{frontmatter}




% \section*{Declaration} 
% The authors declare that they have no known competing financial interests or personal relationships that could have appeared to
% influence the work reported in this paper, that there are no conflicts of interest involved in the preparation and submission of this manuscript, and that ethical standards were upheld in the preparation of this work.

% \section*{Acknowledgements}
% This research/project is supported by the NUS Sustainable and
% Green Finance Institute (SGFIN), NUS Asian Institute of Digital Finance
% (AIDF), Ministry of Education, Singapore under its MOE Academic
% Research Fund Tier 2 (STEM RIE2025 Award MOE-T2EP20123-0005),
% MOE Tier 2 Award (MOE-T2EP50221-0006: ‘‘Prediction-to-Mitigation
% with Digital Twins of the Earth’s Weather’’), MOE Tier 1 Award
% (MOE-T2EP50221-0028: ‘‘Discipline-Informed Neural Networks for Interpretable Time-Series Discovery’’), and by the RIE2025 Industry
% Alignment Fund – Industry Collaboration Projects (IAF-ICP) (Award
% I2301E0026), administered by A*STAR, as well as supported by Alibaba Group and NTU Singapore.

% \begin{linenumbers}
% \newpage
\section{Introduction}\label{sec:intro}

%%% problem; background intro
Amid growing demands for corporate sustainability—companies' commitment to reducing environmental impact, promoting social responsibility, and upholding ethical governance~\citep{ong2024xnlpsusanalysis}—regulators and investors are increasingly focused on corporate sustainability analysis~\citep{cenci2023corporatesusimpt}. This involves evaluating a company's progress toward Environmental, Social, Governance (ESG) goals, otherwise known as ESG assessment, with company-disclosed sustainability reports often used for analysis~\citep{nguyen2020keysource}. However, the complexity and sheer volume~\citep{meuer2020naturecorporatesus} of these reports typically render manual analysis impractical~\citep{ni2023chatreport}. To overcome this challenge, NLP-driven ESG topic analysis — the automatic identification of ESG topics within sustainability reports — has been widely implemented. By enabling the automated and structured analysis of sustainability reports, ESG topic analysis provides regulators and investors key insights on a corporation's sustainability performance, such as extent of alignment with environmental objectives~\citep{WANG2020sdg}, future orientation of sustainability activities~\citep{HEICHL2023textmining}, among others. These insights are useful for a variety of applications, such as sustainable portfolio construction~\citep{henriksson2019integrating} and regulatory auditing~\citep{ferjanvcivc2024textualanalysis}, which significantly underpin the field of sustainable finance.

% definition of topic analysis, contextualised within sustainability
However, despite advancements in NLP, existing methods struggle to meet the specialised demands of ESG topic analysis, limiting their effectiveness for this task. ESG topic analysis often aims to capture ESG-related content that is both relevant for ESG assessments and aligned with established ESG frameworks~\citep{ong2024xnlpsusanalysis}, such as SASB~\citep{eng2022sasb} and GRI~\citep{machado2021gri}. To accomplish this, commonly used NLP techniques, including rule-based dictionary methods~\citep{baier2020dict} and unsupervised topic modeling~\citep{blei2003lda}, may fall short as they lack task-specific design. We identify three key shortcomings below:
\begin{itemize} 
\item \textbf{Irrelevance}: 
Existing topic modeling methods, while excellent at producing topic terms without supervision~\citep{abdelrazek2023topicsurvey2}, cannot be tuned to capture ESG content. Therefore, captured topic terms may not be relevant to ESG, leading to topics irrelevant for ESG analysis. Additionally, sustainability dictionaries that are popularly adapted for ESG topic analysis - i.e Baier~\citep{ignatov2023esgdict}, comprise a considerable portion of non-ESG terms.
\item \textbf{Immateriality}: Stakeholders often prioritise content that explicitly conveys the ESG efforts or actions undertaken by corporations, as this content has greater \textit{materiality} or significance for ESG performance~\citep{beske2020sustainabilityactionsmaterial}. However, both rule-based dictionaries and topic models do not focus on extracting topic terms that capture ESG-oriented actions. To better align with stakeholder interests, topic terms should be refined to reflect these actions.
\item \textbf{Limited Organisation}: Due to its unsupervised nature, topic modeling may generate topics that are not organised according to ESG frameworks. Consequently, even if ESG-related topics emerge, their lack of organisation means that they fail to comprehensively capture all dimensions of ESG. This incomplete coverage can be problematic for sustainability stakeholders, who rely on structured ESG frameworks to ensure that all relevant ESG aspects are analysed~\citep{malesios2021sustainabilityframework1,eslami2021analysingframework2}. While sustainability dictionaries offer a more organised approach~\citep{HEICHL2023textmining}, their limited size (fewer than 800 terms) constrains their coverage, potentially restricting the range of ESG topics identified in practice.
\end{itemize}

Addressing the above limitations requires an ESG topic analysis approach that explicitly captures relevant, material, and well-structured ESG content. To this end, we explore the following research questions:

\begin{itemize}
\item \textbf{RQ1}: How can a topic analysis method be designed to mitigate the limitations of existing NLP techniques in capturing relevant and material ESG content according to structured ESG frameworks?

\item \textbf{RQ2}: What criteria and evaluation methods are appropriate for assessing the effectiveness of ESG topic analysis?

\item \textbf{RQ3}: To what extent does a method specifically designed for ESG topic analysis outperform existing NLP approaches for this task?
\end{itemize}

To contribute to these research questions, we develop ESGSenticNet, an NLP tool specifically designed to facilitate ESG topic analysis. We then evaluate it extensively against state-of-the-art NLP methods, using specialised evaluation metrics tailored for ESG topic analysis.

ESGSenticNet is a publicly available knowledge base comprising 44k sustainability knowledge triples from 23k unique concepts. While ESGSenticNet itself is a knowledge base, it functions as a structured lexicon for rule-based topic analysis. To elaborate, each triple within ESGSenticNet follows the format (\textit{concept, relation, ESG topic})— i.e. (`minimise paper waste', supports, `waste management'). During topic analysis, the concepts serve as topic terms, which can be matched within sustainability reports to indicate the presence of specific ESG topics. Unlike existing topic analysis methods, ESGSenticNet explicitly addresses \textbf{irrelevance, immateriality} and \textbf{limited organisation}, through its construction from a neurosymbolic framework. This framework integrates symbolic linguistic rules and sub-symbolic deep-learning to ensure ESGSenticNet's triples comprise relevant and material ESG concepts that are effectively organised according to structured ESG topics.

To reduce \textbf{irrelevance}, the symbolic component applies sustainability-specific linguistic patterns to extract ESG-related concepts, thereby minimising the occurrence of irrelevant topic terms in ESG topic analysis. To address \textbf{immateriality}, the applied linguistic patterns also prioritise extracting concepts that explicitly describe ESG actions, making topic terms more meaningful for stakeholders seeking actionable ESG insights. To ensure \textbf{organisation}, the sub-symbolic component leverages GPT-4o inference, supported with a graph-based semi-supervised method, to categorise concepts according to structured ESG topics under our constructed taxonomy. Put together, the neurosymbolic framework yields triples in the form of (\textit{concept, relation, ESG topic}), and ensures that ESGSenticNet addresses the shortcomings of existing NLP for ESG topic analysis.

We evaluate ESGSenticNet’s effectiveness in ESG topic analysis by benchmarking it against state-of-the-art NLP algorithms commonly used for this task. Given the unique requirements of ESG topic analysis—capturing ESG content material for assessments, and aligned with established ESG frameworks—we introduce specialised evaluation metrics, namely (1) \textit{ESG-relatedness}: The proportion of all topic terms relevant to ESG; (2) \textit{ESG-action orientation}: The proportion of all topic terms focused on ESG actions, representing their material significance in sustainability assessments; (3) \textit{Unique ESG topic terms}: The number of unique ESG terms, which highlights the comprehensiveness and diversity of ESG content captured. (1) and (2) directly addresses the core challenges of \textbf{irrelevance} and \textbf{immateriality} respectively, and (3) is related to the method’s practical utility in real-world deployments.

We show that ESGSenticNet significantly outperforms existing baselines on these three metrics, achieving 76\% ESG-relatedness  (+26\% improvement) and 84\% ESG action-orientation (+34\% improvement), while capturing a high number of unique ESG terms 359 under the \textit{exact} implementation; 2555 under the \textit{flexible} implementation). 
Additionally, human evaluation confirms that ESGSenticNet correctly categorises concepts under their respective ESG topics with 86\%+ accuracy. 
As a result, during topic analysis, the topic terms matched from ESGSenticNet align with their associated ESG topics with considerable reliability. 
Finally, as a lexical method, ESGSenticNet requires no training or technical expertise, making it accessible to stakeholders across diverse backgrounds. To underscore the significance of these results for ESG topic analysis, we relate them back to the three core challenges—\textbf{irrelevance}, \textbf{immateriality}, and \textbf{limited organisation}—as well as the practical concerns of \textbf{comprehensiveness \& diversity} and \textbf{computational complexity}. While the three core challenges have been previously defined, \textbf{comprehensiveness \& diversity} captures the extent and variety of ESG content each method is able to extract, and \textbf{computational complexity} refers to the ease of implementation and resource demands of each method. Table~\ref{tab:limitations_comparison} qualitatively compares ESGSenticNet, topic models, and existing lexicons in terms of how well they address each of these five dimensions. These assessments are informed by our experimental results—ESG-relatedness corresponds to \textbf{irrelevance}, ESG action-orientation to \textbf{immateriality}, and the number of unique ESG terms to \textbf{comprehensiveness \& diversity}—as well as qualitative evaluations for \textbf{limited organisation} and \textbf{computational complexity}.

\begin{table}[H]
\centering
\small % Reduce font size for better fit
\rowcolors{2}{white}{blue!5}
\begin{tabularx}{\linewidth}{p{1.8cm} X p{1.9cm} p{1.8cm} X X}
\rowcolor{blue!20}
\textbf{Method} & \textbf{Irrelevance} & \textbf{Immateriality} & \textbf{Lim-Org} & \textbf{Compre-Div} & \textbf{Complexity} \\
\textbf{Topic \linebreak Models} & Extracted terms are largely irrelevant to ESG topics & Extracted terms rarely reflect concrete ESG action-orientation, reflecting immateriality & Unstructured and not aligned with ESG frameworks & Specific models capture a high amount of ESG terms, although majority of extracted terms are not ESG related & High: requires extensive hyperparameter tuning and model selection \\
\textbf{Existing \linebreak Lexicons} & More relevant than topic models, but still often include unrelated terms & Partially captures but still largely omits ESG action-oriented terms & Structured by a single ESG taxonomy & Captures a small to moderate amount of ESG terms & Low: no training or tuning required \\
\rowcolor{blue!5}
\textbf{ESGSenticNet} & Extracted terms are consistently ESG-relevant & Strong focus on action-oriented, material ESG terms & Structured by a hierarchical ESG taxonomy aligned with real-world reports, with high concept-topic relation accuracy. & Captures a broad and diverse amount of ESG terms & Low: no training or tuning required \\
\end{tabularx}
\caption{Qualitative comparison of ESG text analysis methods across five key dimensions. Lim-Org stands for \textbf{limited organisation}, Compre-Div for \textbf{comprehensiveness \& diversity}, Complexity for \textbf{computational complexity}. Each assessment reflects evidence from our experimental findings where applicable — for example, high irrelevance corresponds to low ESG-relatedness scores, high immateriality to low ESG action-orientation scores, and limited comprehensiveness to low unique ESG terms captured. Limited organisation and computational complexity are assessed through qualitative evaluations.}
\end{table}\label{tab:limitations_comparison}

The findings position ESGSenticNet as a tool that significantly enhances ESG topic analysis, offering clear advantages over existing approaches across all five dimensions.

The rest of our paper is arranged as follows. We begin with a literature review (Section~\ref{sec:litreview}). Then, we explain the preliminary steps (Section~\ref{sec:preliminaries}) prior to constructing the knowledge base, which involves describing the data utilised in this study (Section~\ref{sec:data}), and defining the knowledge base's organisational framework. The latter involves establishing a hierarchical ESG taxonomy that defines the topics in our knowledge base (sections~\ref{sec:taxonomy},~\ref{sec:taxsection}) and specifying the relations and rules that govern its structure (Section~\ref{sec:rules}). 

After outlining the prelimnary steps, we delve into our neurosymbolic framework for constructing the knowledge triples within ESGSenticNet (Section~\ref{sec:esgsenticnet_entire_construction}). This involves the extraction of concepts via ESG concept parser (Section~\ref{sect:esgconceptparser}), quality control—processing the extracted concepts to ensure their coherence (Section~\ref{sec:processingconcepts}), followed by the labelling process for classifying these concepts according to our taxonomy topics (Section~\ref{sec:labellingprocess}). This labelling process involves leveraging GPT-4o inference, supported by a graph-based semi-supervised method (sections~\ref{sec:semgraphconstruct},~\ref{sec:seed select},~\ref{sec:seed annotation},~\ref{sec:lpa}).
% The development process for ESGSenticNet is summarised by Figure 1, showcasing the key steps for concept parsing and taxonomy derivation, and how they are intertwined with the labelling process (1-4). 

Our neurosymbolic framework (Section~\ref{sec:esgsenticnet_entire_construction}) results in the construction of ESGSenticNet, which we extensively evaluate through experiments on topic analysis, with the results and implications discussed (Section~\ref{sec:topicevaluation}).

\section{Results}\label{sec:topicevaluation}
% why this is important
To evaluate the effectiveness of ESGSenticNet against existing NLP methods, we perform topic analysis on the evaluation set (described in Section~\ref{sec:data}). This set comprises 319 sustainability reports from SGX companies, and the reports were unused during the development of ESGSenticNet. In the following, we discuss the evaluation metrics (Section~\ref{sec:evalmetrics}) and experimental settings (Section~\ref{sec:experiments}) utilised during this experiment, before discussing the experimental results (Section~\ref{sec:results}). Thereafter, we further evaluate the accuracy of ESGSenticNet's triples through human evaluation (Section~\ref{sec:evaluation}), before discussing a case study of ESGSenticNet's application (Section~\ref{sec:casestudy}) and the implications of our ESGSenticNet study (Section~\ref{sec:implications}). We finish this section with additional ablation studies (Section~\ref{sec:ablation}) to improve the robustness of our findings.

% In the following, we detail the evaluation metrics used in our experiment (Section~\ref{sec:evalmetrics}), prior to describing the baseline NLP methods and their experimental settings (Section~\ref{sec:experiments}).

\subsection{Evaluation Metrics}\label{sec:evalmetrics}

% Kang & 21* & \underline{0.75}* & \underline{0.57}* & 21* & \underline{0.75}* & \underline{0.57}* & 21* & \underline{0.75}* & \underline{0.57}* \\

As discussed earlier (Section~\ref{sec:intro}), ESG topic analysis has distinct requirements as it supports sustainability assessments. It focuses on extracting ESG content that is significant for these assessments, and organised within established ESG frameworks. Therefore, while topic analysis metrics—specifically pertaining to topic models (i.e. perplexity, PMI)~\citep{abdelrazek2023topicsurvey2}—have been explored, ESG topic analysis require specialised metrics. Therefore, methods are evaluated according to the following:
\begin{itemize}
    \item \textbf{ESG relatedness}: The proportion of all topic terms that are ESG-related,  highlighting the extent to which yielded topics are relevant to ESG and therefore ESG analysis.
    \item \textbf{ESG action orientation}: The proportion of all topic terms that express an action taken toward ESG, or in other words, an ESG effort. This indicates whether the terms are able to express information \textit{material} to ESG assessments, and of interest to stakeholders. 
    \item \textbf{Unique ESG terms}: The number of unique ESG topic terms, indicating the comprehensiveness and diversity of ESG content captured.
\end{itemize}

% \begin{itemize}[leftmargin=*]
%     \item \textit{ESG relatedness}. The proportion of all topic terms that are ESG-related,  highlighting the extent to which yielded topics are relevant to sustainability and therefore sustainability analysis. 
%     \item \textit{Unique ESG terms}. The number of unique ESG topic terms, indicating the comprehensiveness and diversity of ESG content captured. 
%     \item \textit{ESG action orientation}. The proportion of all topic terms that express an action taken toward ESG, or in other words, a sustainability effort. This is indicative of whether the terms are able to express information \textit{material}, and of interest to stakeholders. 
% \end{itemize}

 % out of all the topic terms, following from~\citep{diengdiversitymetric}.

For the above metrics, GPT-4 is leveraged to classify all topic terms yielded from the models, aligning with the LLM-as-a-judge method from established NLP studies~\citep{zheng2023llmjudging}. GPT-4 distinguishes the following -- 1) whether each topic term is ESG-related, 2) whether each topic term constitutes an action toward improving ESG. Thereafter, we aggregate the GPT-4 classification results to produce scores for \textit{ESG relation, unique ESG terms, ESG action}. We utilise GPT-4 in our work given that it offers a scalable, automated, and reproducible approach to evaluating ESG topic terms~\citep{chiang2023llmhuman} — a task that would be prohibitively costly and impractical to perform manually, given the tens of thousands of topic terms involved. At the same time, we recognise the potential limitations of LLM-based evaluation~\citep{zheng2023llmjudging}. Therefore, to ensure robustness, we validate the use of GPT-4 in our evaluation through an GPT4-human agreement study (Section~\ref{sec:agreement}). Full experimental details for topic analysis evaluation are described in Section~\ref{sec:experiments}, with the prompts provided in Figure~\ref{prompt:esgrel} (ESG relatedness), and Figure~\ref{prompt:esgaction} (ESG action orientation). 
\begin{singlespace}
\begin{figure}[H]
    \centering
    \small
    \begin{tcolorbox}[colframe=black, colback=white, boxrule=0.8pt, width=\textwidth]
    \textbf{[Task Description]} \\
    As a sustainability expert, your task is to classify terms based on their relevance to Environmental, Social, and Governance (ESG) factors. \\[5pt]
    
    \textbf{[Response Instructions]} \\
    Evaluate each term based on its content to determine its most appropriate topic. If a term is explicitly related to ESG, assign the label `ESG'. If a term is not related to ESG or the relationship is vague, assign the label `non-ESG'. Ensure all responses strictly adhere to the definitions and format provided. \\[5pt]
    
    \textbf{[Response Format]} \\
    Fill in your response in the triple backticks below using the following format: \\
    \texttt{Input: <term>} \\
    \texttt{Response: <ESG or non-ESG>}
    \end{tcolorbox}
    \caption{Prompt for evaluating ESG relatedness}
    \label{prompt:esgrel}
\end{figure}

\begin{figure}[H]
    \centering
    \small
    \begin{tcolorbox}[colframe=black, colback=white, boxrule=0.8pt, width=\textwidth]
    \textbf{[Task Description]} \\
    As a sustainability expert, your task is to classify terms based on whether they express an action taken toward ESG. \\[5pt]
    
    \textbf{[Response Instructions]} \\
    Evaluate each term based on its content to determine whether it describes an action taken to advance Environmental, Social, and Governance (ESG) goals. If a term describes an action taken to enhance ESG objectives, assign the label `True'. If a term does not describe an action taken toward ESG objectives, assign the label `False'. Ensure all responses strictly adhere to the definitions and format provided. \\[5pt]
    
    \textbf{[Response Format]} \\
    Fill in your response in the triple backticks below using the following format: \\
    \texttt{Input: <term>} \\
    \texttt{Response: <True or False>}
    \end{tcolorbox}
    \caption{Prompt for evaluating ESG action orientation}
    \label{prompt:esgaction}
\end{figure}
\end{singlespace}

% \subsection{Experimental Settings}\label{sec:expt}
\subsection{Experimental Settings}\label{sec:experiments}
For this study, the following baselines were utilised. Topic models that are algebraic -- NMF~\citep{nmflee1999}, probabilistic -- LDA~\citep{blei2003lda}, neural -- ProdLDA~\citep{srivastava2017prodlda}, TSCTM~\citep{wu2022tsctm}, BERTopic~\citep{grootendorst2022bertopic}, ECRTM~\citep{wu2023ecrtm}, and other publicly available ESG dictionaries -- Baier~\citep{baier2020dict} Naiara~\citep{Naiara2024news}. To the best of our knowledge, ESG dictionary methods in the published literature are limited, with Baier and Naiara among the only few works that make their full dictionaries publicly available. 

Algebraic, Probabilistic \& Neural methods are unsupervised, and we run them across a wide range of hyperparameters, to obtain the best results for each method. These parameters include every combination for vocab size$=\{2500, 5000, 7500, 10000, 12500\}$, number of topics$=\{10, 20, 30, 40, 50, 60, 70\}$, number of topic words$=\{5,10,15,20,25,30\}$. These models were deployed through the TopMost package~\citep{wu2023topmost}. 

Dictionary methods comprise terms that are categorised according to ESG topics, with a topic's occurrence computed from the frequency of occurrence of its associated terms. In our experiment, the topic terms of the dictionaries correspond to dictionary terms that appear within the corpora of sustainability reports. 

ESGSenticNet comprises knowledge triples in the format~\textit{(concept, relation, ESG topic)}, delineating how specific concepts related with ESG topics or topics. To enhance semantic specificity and context, we employ three word concepts, exclusively using those that possess the `supports' relation toward a ESG topic. Similar to \textit{dictionary} methods, a topic's occurrence is determined by the frequency of appearance of concepts that support the topic (we explore this further in a case study in section (\ref{sec:casestudy})). The topic terms of ESGSenticNet correspond to these concepts that appear within the sustainability reports corpora. Additionally, we experiment with two configurations for ESGSenticNet:~\textit{Exact} involves exact string matches of each concept,~\textit{Flexible} involves taking advantage of the (verb, noun phrase) arrangement of concepts for matching. To elaborate on this, the verb (1st word) is matched separately from the corresponding noun-phrase (2nd \& 3rd words) within each sentence. Given the example ``We reduce our water consumption'', `reduce' is matched as the verb, and `water consumption' is matched as the noun-phrase, with `reduce water consumption' returned as the matched concept without requiring its exact string match.~\textit{Flexible} enables greater detection of ESG concepts by allowing for semantic variability (see study in Section~\ref{sec:flexible}), while~\textit{Exact} is useful for detecting precise ESG impacts. 

\subsection{Experimental Results}\label{sec:results}
\begin{table}[H]
\resizebox{\textwidth}{!}{%
\setlength\tabcolsep{2.3pt}
\begin{tabular}{lllllllllllllllll}
\hline \hline
 \multirow{2}{*}{Baseline} & \multicolumn{3}{c}{\textbf{Best ESG-unique}} & \multicolumn{3}{c}{\textbf{Best ESG-rel}}  & \multicolumn{3}{c}{\textbf{Best ESG-act}} \\
 & ESG-unique & ESG-rel & ESG-act & ESG-unique & ESG-rel & ESG-act & ESG-unique & ESG-rel & ESG-act & \\ \hline

NMF & 332* & 0.28* & 0.08* & 3 & 0.42 & 0.00 & \underline{332}* & 0.28* & 0.08* \\
LDA & 23* & 0.31* & 0.01* & 3 & 0.36 & 0.00 & 23* & 0.31* & 0.01*\\
ProdLDA & 334* & 0.26* & 0.07* & 17 & 0.38 & 0.10 & 13 & 0.38 & 0.16 \\
BERTopic & 76 & 0.33 & 0.05 & 13 & 0.40 & 0.04 & 13 & 0.35 & 0.05\\
TSCTM & 359 & 0.15 & 0.06 & 11 & 0.32 & 0.06 & 23 & 0.31 & 0.15\\
ECRTM & \underline{440} & 0.18 & 0.09 & 23 & 0.46 & 0.08 & 182 & 0.31 & 0.15 \\ \hline
Baier & 212* & 0.49* & 0.29* & \underline{212}* & 0.49* & 0.29* & 212* & 0.49* & 0.29* \\
Naiara & 1* & \underline{0.50}* & \underline{0.50}* & 1* & \underline{0.50}* & \underline{0.50}* & 1* & \underline{0.50}* & \underline{0.50}* \\
\hline
ESGSenticNet (exact) & 359* & 0.76* & \textbf{0.84}* & 359* & 0.76* & \textbf{0.84}* & 359* & 0.76* & \textbf{0.84}* \\
ESGSenticNet (flexible) & \textbf{2555}* & \textbf{0.79}* & 0.81* & \textbf{2555}* & \textbf{0.79}* & 0.81* & \textbf{2555}* & \textbf{0.79}* & 0.81*\\
\hline
\end{tabular}%
}
\caption{Evaluation of Topic Analysis Methods, through the metrics of unique ESG terms (ESG-unique), ESG relatedness (ESG-rel), and ESG action orientation (ESG-act). Best results are marked in bold and the best baseline results are underlined. (*) indicates the results from the same run (i.e. same parameters). Best ESG-unique, Best ESG-rel, and Best ESG-act represent the top-performing results from the respective runs of the tested methods.}
\label{tab:results}
\end{table}

\begin{figure}[H]
    \centering
    \hspace*{-1.5cm}
    \includegraphics[width=0.8\textwidth]{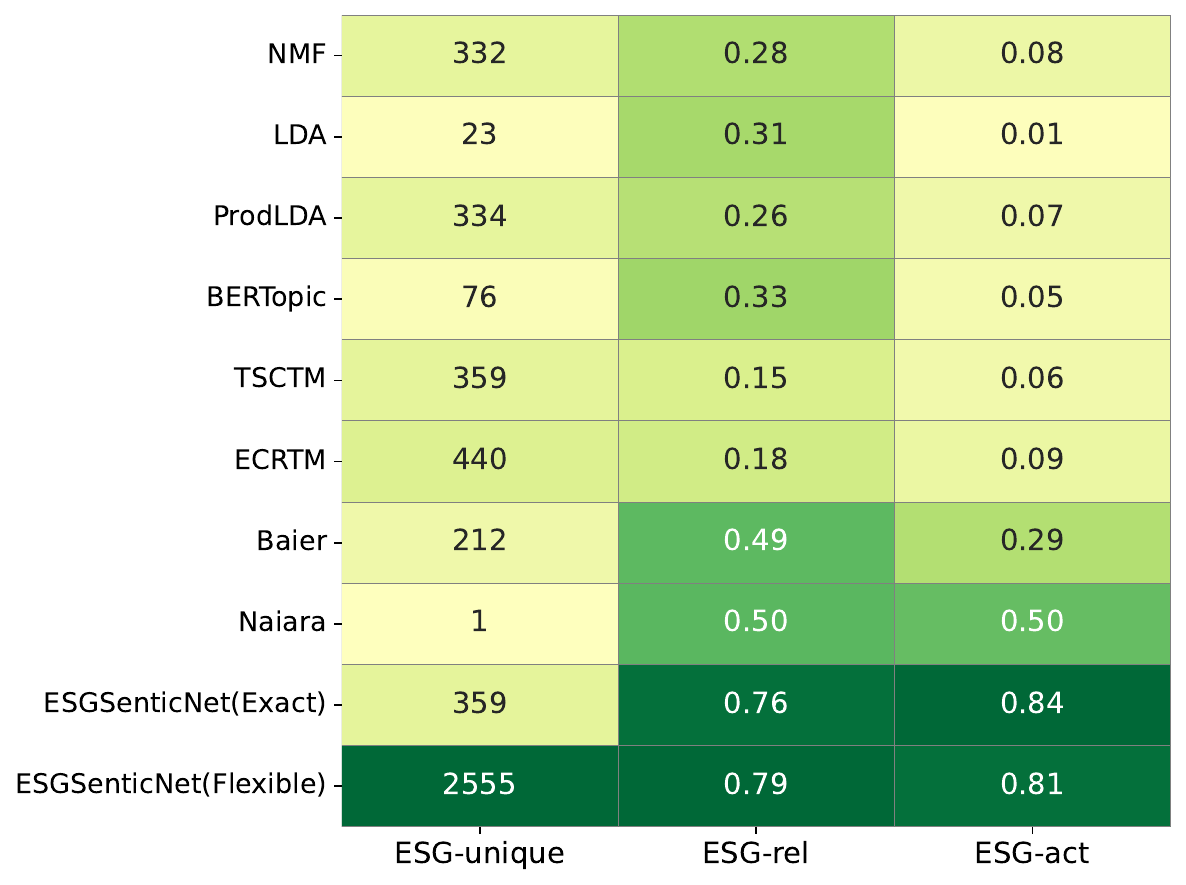}
    \caption{Heatmap highlighting performance of different topic analysis methods on the metrics of number of unique ESG terms (ESG-unique), ESG relatedness (ESG-rel), ESG action orientation (ESG-act). ESGSenticNet(Exact) refers to ESGSenticNet deployed by \textit{Exact} matching, while ESGSenticNet(Flexible) refers to ESGSenticNet deployed by \textit{Flexible} matching, as explained in Section 5.2. Darker green indicates better performance, lighter green indicates poorer performance.}
    \label{fig:esg_metrics_performance}
\end{figure}

Our experimental results (Table~\ref{tab:results}) indicate that ESGSenticNet scores higher than the other topic analysis methods on \textit{ESG relatedness} and \textit{ESG action orientation} by at least 26\% and 31\% respectively. Additionally, both ESGSenticNet (exact) and ESGSenticNet (flexible) extract a high number of unique ESG terms, with ESGSenticNet (flexible) scoring the highest amongst all methods tested. Although requiring limited computational resources as a lexical method, ESGSenticNet produces significantly higher performance compared to the more expensive unsupervised topic modeling methods. This is despite how the topic models were exhaustively ran on numerous permutations of parameters (Section~\ref{sec:experiments}) that required extensive computational costs.

\begin{table}[H]
\centering
\begin{tabular}{|l|l|p{7.5cm}|}
\hline
\textbf{Method} & \textbf{Topic} & \textbf{Example Topic Words} \\ 
\hline
TSCTM & Topic 1 & offshore, baker, vessels, oil, cho, vessel, chartering, ship, sea charter \\
      & Topic 2 & rsp, venture, johor, black, sites \\
      & Topic 3 & controlling, provision, shall, period, accounting, mainly, listing, media, entity, time \\ \hline
ECRTM & Topic 1 & minerals, healthcare, disa, teh, bhd holdings, agm, sdn, heatec, sri \\
      & Topic 2 & banks, grand employees, duty, palm \\
      & Topic 3 & disa, shares, theft share, anniversary, lau, purchase, options, june cancellation \\ \hline
ProdLDA & Topic 1 & financial, value, non assets, december \\
        & Topic 2 & loss, company, board, statements, directors \\
        & Topic 3 & greenhouse, mountain, aspires, gresb, honeyness, feel, strategy, board, candid, phenomenon, conforms, performance, cubic, cabling, unoccupied \\ \hline
NMF & Topic 1 & duly, assets, transparency, value, priority \\
    & Topic 2 & property, fair, rate, million, new \\
    & Topic 3 & esg, administration, confirmed, quarter, save, present, businesses, employees, workplace, flow, focusing, ceo, default, meetings, termination \\ \hline
\textbf{ESGSenticNet} & Topic 1 & abide local law, achieve regulatory compliance, fulfill regulatory requirement, follow international law, comply environmental rule \\
             & Topic 2 & increase employee satisfaction, improve staff engagement, invest career development, maintain employee safety \\
             & Topic 3 & minimise carbon footprint, monitor ghg emission, reduce environmental pollution, lower air emission \\ 
             \hline
\end{tabular}
\caption{Examples of topic words for topic models with the highest scores on the number of unique ESG terms.}
\label{tab:expt_samples}
\end{table}

Unsupervised topic models such as TSCTM, ECRTM, ProdLDA, NMF can extract a high number of unique ESG terms ($>300$). Yet, for the same runs, the concentration or prevalence of ESG terms within each topic is modest, as highlighted by their limited scores in \textit{ESG relatedness} ($<0.30$). Therefore, despite capturing diverse ESG topic terms, these methods fall short in generating topics that are richly centered on ESG themes. From Table~\ref{tab:expt_samples}, we highlight examples of the topics generated by these methods in comparison to ESGSenticNet, highlighting how most of the topics generated by these topic models are not ESG-centric. This limitation is critical in ESG topic analysis, given that the primary goal is to derive insights from topics that are not only inclusive of ESG terms but are also substantially focused on ESG issues. 

We also compare the number of unique ESG-related terms extracted by methods that construct structured ESG lexicons from existing sustainability corpora—specifically Baier and Naiara, as this serves as an indicator of lexical coverage and diversity. ESGSenticNet yields 359 unique terms in its exact matching configuration and 2,555 under its more flexible variant, compared to 212 for Baier and just 1 for Naiara. These differences suggest that ESGSenticNet captures a broader and more diverse set of ESG terms from the test corpora, reflecting a more extensive and diverse collection of ESG lexicons (or ESG concepts) relative to the Baier and Naiara lexicons. This outcome reflects the advantages of our neuro-symbolic framework—discussed in Section~\ref{sec:litrev_lexicon}—over manual or anchor-based lexicon construction methods, as it enables scalable and systematic identification of ESG concepts through automated concept parsing, GPT-4o-based annotation, diversity-aware seed selection, and label propagation.

Moreover, topic model runs that yield the highest \textit{ESG relatedness} are only able to yield few unique ESG-terms ($<25$). Furthermore, for these topic model configurations, as well as for other dictionaries such as Baier and Naiara, \textit{ESG relatedness} is still $\leq 0.50$. This suggests that yielded topics are still not predominantly centred around ESG themes.

Finally, most of the tested baseline methods score modestly for \textit{ESG action orientation}, Naiara noticeably higher. Yet, Naiara is only able to derive $1$ unique ESG-term, which limits the significance of its analysis. These results suggest that the baseline methods may not fully capture information that is \textit{material} to stakeholders, particularly in terms of conveying ESG actions and efforts. 

% Ultimately, we hope our work encourages further exploration and refinement of NLP in this field.
  
% % neurosymbolic 
% Beyond lexical analysis, ESGSenticNet aims to support the development of NLP algorithms that are well-adapted and specialised for the sustainability domain. Recognising that sustainability language is distinct and often not fully captured by general pre-trained models~\citep{gururangan2020domainadapt}, ESGSenticNet can enrich these models with deep domain knowledge, similar to~\citep{cambria2024senticnet8, liang2022aspectGCN, zhang2023neuro, nlu}. 

%Sustainability encompasses specific and distinct language that general pre-trained models may not fully understand~\citep{gururangan2020domainadapt}. For instance, while `reduce consumption' may imply resource stewardship (enhancing resource efficiency) within sustainability, it can carry a different implication in other domains like finance (reduced consumption within an economy). Without a significant grasp of sustainability jargon, the capabilities of language models within this domain will be limited. To this end, ESGSenticNet contains knowledge that can potentially be embedded in language models, similar to~\citep{cambria2024senticnet8, liang2022aspectGCN, zhang2023neuro, nlu}, thereby enhancing natural language understanding within the sustainability domain.

\subsection{ESGSenticNet Human Evaluation}\label{sec:evaluation}

In this section, we evaluate ESGSenticNet’s effectiveness in categorising concepts under the correct topics. This ensures that during topic analysis, the topic terms matched from ESGSenticNet reliably correspond to their intended ESG topics. To assess this, human annotators determine the accuracy of relations between concepts and topics. 500 concepts are randomly selected from our knowledge base. A concept is found within a knowledge triple, possessing a relation with a topic in the format of (\textit{concept}, \textit{relation}, \textit{ESG topic}). A correct \textit{relation} represents that the \textit{concept} is accurately classified under the \textit{ESG topic}. The accuracy of the relations are evaluated separately for each topic type—\textit{pillar, broad, cross-broad, sub, cross-sub}, with each of them containing different ESG topics as described in Table \ref{tab:category_type_descriptions}. Human annotators adhere to the following instructions (\ref{sec:humanannotate} provides more details on our annotators and annotation scheme): 

\begin{singlespace}
\noindent\fbox{%
  \begin{minipage}{\dimexpr\columnwidth-2\fboxsep-2\fboxrule}
    \small The table below presents various concepts along with their polarities and respective relations toward ESG topics. Each concept may be associated with multiple relations across different ESG topics. These topics are classified into five topic types: pillar, broad, sub, cross-broad, and cross-sub. If a concept has no relevant association with a topic, it is marked as `not applicable'. Your task is to evaluate whether each concept is correctly related to its corresponding topics. For each topic, determine if the relation is the most appropriate compared to the other topics within the same topic type. Mark `TRUE' if the relation and topic are correct, and `FALSE' if they are incorrect. Additionally, assess the accuracy of the polarity of each concept, and mark `TRUE' if it is correct and `FALSE' if it is incorrect.
  \end{minipage}
}
\end{singlespace}

\begin{singlespace}

\begin{table}[h]
\centering
\begin{tabular}{|l|l|}
\hline
\textbf{Topic Type} & \textbf{Accuracy} \\ \hline
Pillar & 95.0\% \\ \hline
Broad & 89.3\% \\ \hline
Cross-Broad & 86.1\%\\ \hline
Sub & 87.8\%\\ \hline
Cross-Sub & 90.1\%\\ \hline
\end{tabular}
\caption{Accuracy of ESGSenticNet for the relations across all topic types.}
\label{tab:esgsenticnet_accuracy}
\end{table}
\end{singlespace}

Results from the human evaluation study highlight that ESGSenticNet retains a high degree of accuracy at 86+\% for relations across all topic types, as shown in Table~\ref{tab:esgsenticnet_accuracy}. These findings suggest that ESGSenticNet has reasonable accuracy in categorising concepts under the correct topics. Consequently, the topic terms matched from ESGSenticNet during topic analysis corresponds to its related ESG topic with considerable reliability.

\section{Discussion}

\subsection{Case Study - Insights offered by ESGSenticNet}\label{sec:casestudy}

\begin{figure}[H]
    \centering
    \includegraphics[width=\textwidth]{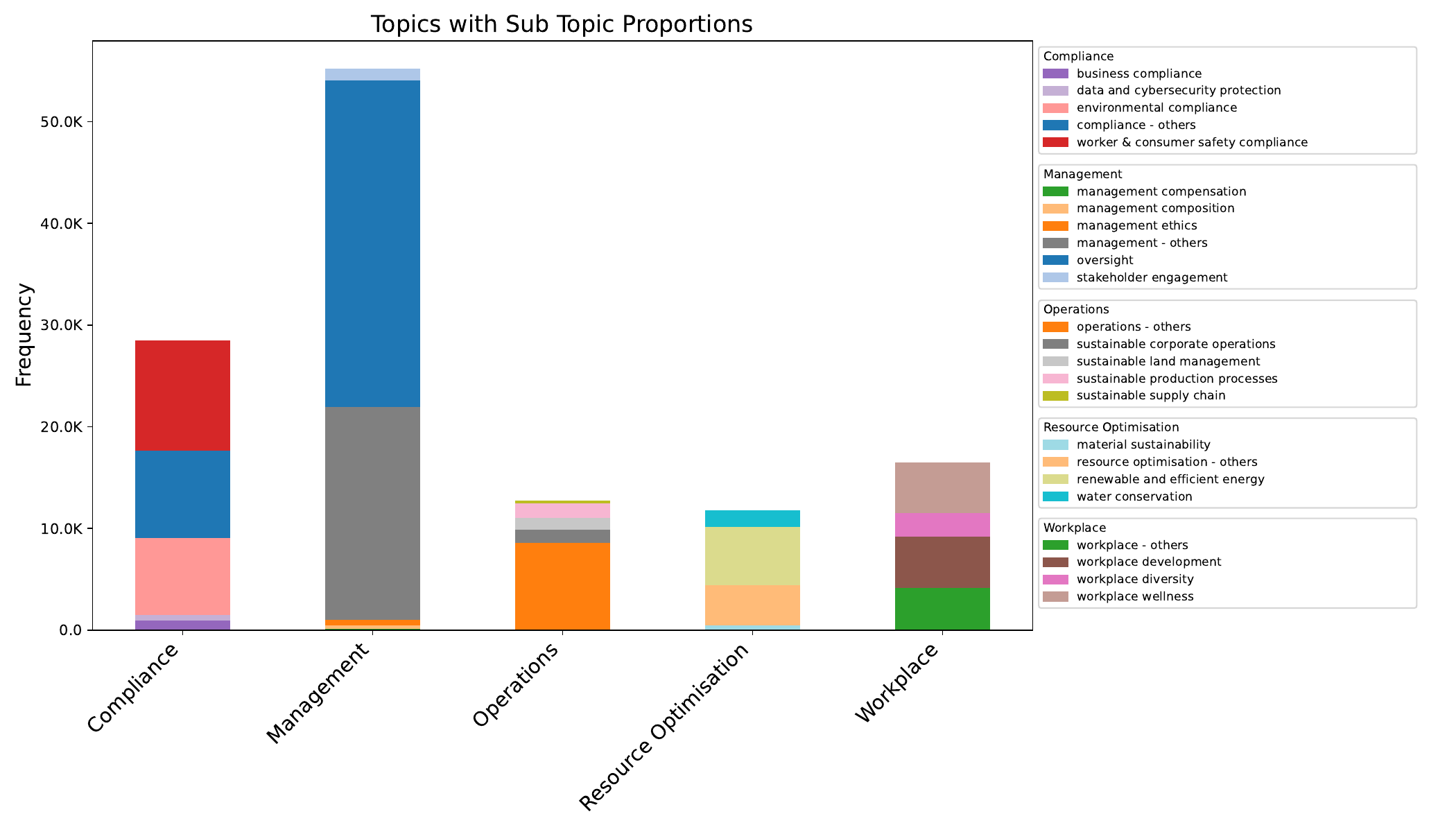}
     \caption{Frequency of ESGSenticNet topics from the~\textit{flexible} run}
    \label{fig:clustered_chart_case_study}
\end{figure}

\begin{figure}[H]
    \centering
    \includegraphics[width=\textwidth]{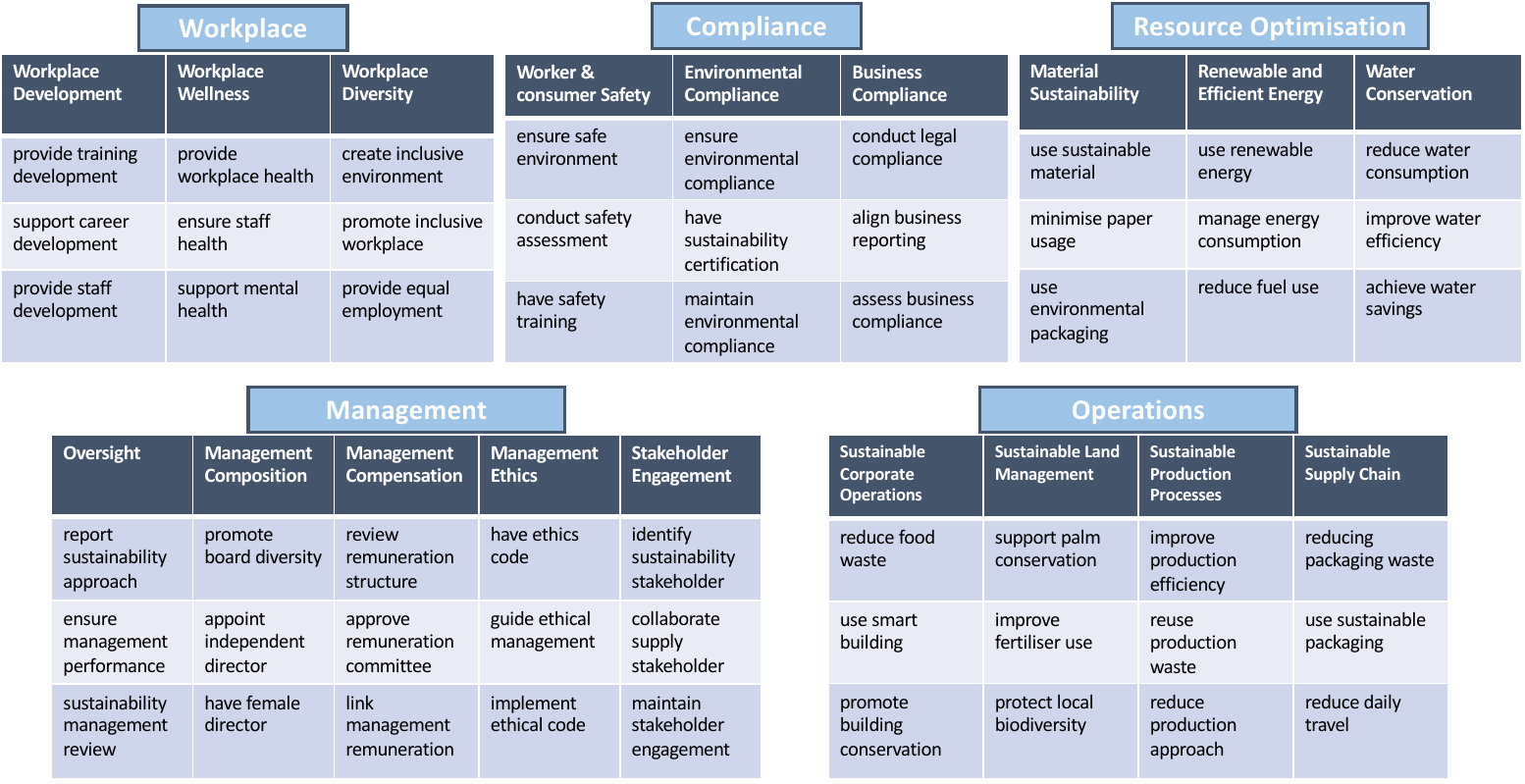}
    \caption{Top ESGSenticNet concepts from the~\textit{flexible} run, in \textit{support} of the sub topics (dark blue), and their corresponding broad topic (light blue)}
    \label{fig:topic_concepts}
\end{figure}

The topic analysis conducted by ESGSenticNet in Section~\ref{sec:topicevaluation} is visualised with the  Figure~\ref{fig:clustered_chart_case_study} \& Figure~\ref{fig:topic_concepts}, which represent different ways to analyse the text corpora. Figure~\ref{fig:clustered_chart_case_study} provides a higher level breakdown of the topics present in the corpora. It highlights the top 5 frequently occurring broad ESG topics present in the corpora, as well as the proportion of their constituent sub topics, in accordance with ESGSenticNet's hierarchical taxonomy. The topic frequencies are denoted by the occurrence of matched concepts that \textit{`support'} a topic, leveraging ESGSenticNet's predefined relations. Separately, Figure~\ref{fig:topic_concepts} provides a more granular analysis of the corpora, highlighting matched concepts that explicitly convey ESG actions, thereby communicating important sustainability information. Figure~\ref{fig:topic_concepts} displays the top occurring concepts that \textit{support} the topics shown in Figure~\ref{fig:clustered_chart_case_study}. Put together, Figure~\ref{fig:clustered_chart_case_study} and Figure~\ref{fig:topic_concepts} highlight different methods to analyse a sustainability corpus, which can be utilised depending on a stakeholder's interest and preference. 

While the case study presents insights derived from a test corpus of 319 SGX sustainability reports, the same form of topical analysis can be applied across different corpora of sustainability reports in various real-world scenarios. We detail examples of these in the following (to qualify, real-world use cases may extend beyond those listed):
\begin{itemize}
\item Analysts for ESG funds can leverage ESGSenticNet's topical insights to evaluate the sustainability of corporations, which is useful for sustainable portfolio construction~\citep{henriksson2019integrating}. In this context, where timely, transparent, and interpretable assessments are essential, using an on-demand and open-source tool like ESGSenticNet for rule-based topic analysis is advantageous over relying on ESG scores, which are often obscure, inconsistent, and infrequently updated~\citep{berg2022aggregate, gyonyorova2023esg}.
\item Regulators can utilise ESGSenticNet to perform sector-wide audits and benchmark companies' sustainability disclosures against each other. For instance, by applying ESGSenticNet across disclosures from firms within a particular industry, regulators can quickly identify under-reporting on material ESG issues (e.g., labour safety in the construction sector, emissions control in the logistics sector)~\citep{ferjanvcivc2024textualanalysis}. This capability enables data-driven supervision and targeted follow-ups. Moreover, as ESGSenticNet operates as an interpretable, rule-based method, it provides transparency in how disclosures are assessed—crucial for public accountability and trust in regulatory oversight.
\end{itemize}
Additionally, compared to the other ESG topic analysis methods (i.e. existing dictionaries and topic modeling), ESGSenticNet's deployment holds significant advantages in that it extracts -- 1) More \textit{relevant} ESG insights, as demonstrated by the greater proportion of ESG related terms (highest ESG-relatedness score in Table 7). 2) ESG insights with greater \textit{materiality} for stakeholders (highest ESG-action orientation score in Table 7). 3) ESG insights that are \textit{organised} according to established ESG frameworks (i.e. the extracted topic terms from ESGSenticNet encompass ESG topics that are structured according to established ESG themes (e.g., compliance, management), as shown in Figure 12, due to ESGSenticNet's taxonomy centric organisation (Section 3).). These advantages position ESGSenticNet as a topic analysis tool that could potentially be more effective than existing solutions for the aforementioned scenarios.

\subsection{Implications}\label{sec:implications}
Our results indicate that ESGSenticNet outperforms existing methods in capturing a significantly higher proportion of ESG-related and ESG-action oriented topic terms while also identifying a large number of unique ESG terms. This suggests that stakeholders using ESGSenticNet for ESG topic analysis can extract more comprehensive and relevant insights from sustainability reports than existing approaches, thereby enhancing ESG assessments. Crucially, this capability could enable new discoveries in sustainable finance and sustainability compliance research, such as examining how textual patterns in sustainability reports relate to firm value~\citep{wang2024decodingfirmvalue} or sustainability violations~\citep{park2024exploringviolations}. ESGSenticNet facilitates structured information retrieval by identifying ESG topics—organised according to a hierarchical taxonomy—within a report and surfacing the specific concepts that constitute them. This structured output can also support document classification by enabling stakeholders to flag disclosures with distinctive reporting patterns, such as under-reporting in key sustainability areas~\citep{ferjanvcivc2024textualanalysis}, as discussed in Section~\ref{sec:casestudy}. These capabilities collectively support real-world decision-making use cases, including portfolio construction for ESG fund managers~\citep{henriksson2019integrating} and sector-level audits for regulators~\citep{ferjanvcivc2024textualanalysis}, as outlined in Section~\ref{sec:casestudy}.

Moreover, our findings highlight that ESGSenticNet can function effectively as a structured lexicon without requiring additional training or tuning, while producing superior results over expensive state-of-the-art NLP topic models. This can make ESG topic analysis more accessible to stakeholders with limited technical expertise or computational resources. Given the interdisciplinary nature of sustainability research, this accessibility is essential for enabling broader adoption by researchers across different fields, thereby facilitating diverse perspectives and approaches that can drive further discoveries in sustainability research.

Additionally, ESGSenticNet's superior performance over existing methods highlights the effectiveness of our scalable and automated neurosymbolic AI approach for lexicon construction in ESG topic analysis. Given its effectiveness, this framework could also be adapted to expand lexicons in other key sustainability contexts that require topic analysis—such as climate risk disclosures~\citep{moreno2022climaterisk} and environmental policy analysis~\citep{camana2021envpolicyanalysis}—where large-scale lexicon development is constrained by high human annotation costs, making automation a necessity.

Moreover, this scalable and automated framework also allows the knowledge base to be continuously updated at minimal cost, ensuring that the lexicons remain relevant over time. This is particularly important in the ESG domain, where sustainability reporting content is dynamic and constantly evolving~\citep{derqui2020towardsevolving}. As sustainability reports change, maintaining up-to-date lexicons is essential for accurately capturing emerging ESG themes and patterns.

\section{Methods}\label{sec:preliminaries}

In this section, we introduce the data utilised in this study (Section~\ref{sec:data}), before defining how ESGSenticNet's triples are organised, laying the groundwork for their subsequent derivation in Section~\ref{sec:esgsenticnet_entire_construction}. As explained earlier in the introduction  (Section~\ref{sec:intro}), ESGSenticNet comprises knowledge triples in the format of (\textit{concept}, \textit{relation}, \textit{ESG topic}). The ESG topics within these triples are drawn from our hierarchical taxonomy in Section~\ref{sec:taxsection}, whose construction methodology is detailed in Section~\ref{sec:taxonomy}. Additionally, we define the relations between concepts and ESG topics within the triples, along with the rules governing these relations, in Section~\ref{sec:rules}.

\subsection{Data}\label{sec:data}
% provide background to the sustainability standards that SGX reports follow
This study utilises a total of 1,998 sustainability reports from 589 Singapore Exchange (SGX) based companies, in the period of 2015 to 2023 inclusive. We randomly divide all the reports into the construction set (1,679 reports) and the evaluation set (319 reports) with a ratio of 85:15, without any overlap between the two sets. The construction set is utilised for developing ESGSenticNet, enabling taxonomy derivation (Section~\ref{sec:taxonomy}) and knowledge base construction (Section~\ref{sec:esgsenticnet_entire_construction}). The evaluation set is reserved for evaluating ESGSenticNet (Section~\ref{sec:topicevaluation}). All reports are publicly accessible online, and are downloaded in PDF format (via the respective company and company-affiliated websites). Thereafter, we processed the report into text format via the PyPDF2 tool. For reproducibility, we provide the full details on the reports in the supplementary material and instructions on how to access these reports. Specifically, we provide CSV files that specify the companies and the corresponding years of each report. Using these details (company name \& report year), all sustainability reports can be easily searched for and downloaded online. Due to space constraints, we can only provide a sample list of the reports we obtained below in Table~\ref{tab:sample_reports}.

\begin{table}[h]
\small
\centering
\begin{tabular}{|l|c|}
\hline
\textbf{Company} & \textbf{Report Date} \\ \hline
ASIAPHOS LIMITED & 2019-12 \\ \hline
MEGROUP LIMITED & 2020-03 \\ \hline
LHN LIMITED & 2020-09 \\ \hline
CHINA REAL ESTATE GRP LIMITED & 2018-06 \\ \hline
CITIC ENVIROTECH LIMITED & 2018-12 \\ \hline
JAWALA INC. & 2019-07 \\ \hline
BUND CENTER INVESTMENT LIMITED & 2018-12 \\ \hline
SAMURAI 2K AEROSOL LIMITED & 2021-03 \\ \hline
HOR KEW CORPORATION LIMITED & 2019-12 \\ \hline
HALCYON AGRI CORPORATION LIMITED & 2021-12 \\ \hline
PROPNEX LIMITED & 2019-12 \\ \hline
NAM CHEONG LIMITED & 2018-12 \\ \hline
HC SURGICAL SPECIALISTS LIMITED & 2019-05 \\ \hline
GKE CORPORATION LIMITED & 2019-05 \\ \hline
\end{tabular}
\caption{Sample list of SGX sustainability reports provided in the supplementary material.}
\label{tab:sample_reports}
\end{table}

% . We will provide a full list of the reports utilised (i.e. the companies, year of reporting, and URLs) for ESGSenticNet construction and testing upon the completion of the review process. 

% Sustainability Taxonomy Construction, Rules \& Relation Definitions

% To further align taxonomies with sustainability report content, data-driven approaches can be utilised for taxonomy derivation~\citep{alnajjar2024topicmodellingsustainabilitytaxonomy}. 

\subsection{Taxonomy Construction Methodology}\label{sec:taxonomy}
ESG taxonomies, particularly those established by regulatory bodies—i.e. GRI~\citep{machado2021gri}—provide structured frameworks for organising sustainability reports, enabling companies to communicate their ESG performance effectively. However, in practice, firms may not align their reporting content with these frameworks, as the frameworks may not fully capture the complexities of real-world business operations~\citep{ong2024xnlpsusanalysis}. To bridge this gap, we augment existing regulatory taxonomies with a data-driven approach that derives additional ESG topics directly from reporting content, involving BERTopic~\citep{grootendorst2020keybert} and human validation. 
% Specifically, we use BERTopic~\citep{grootendorst2020keybert} to extract key ESG topics (in the form of topic words) from the construction set of 1679 sustainability reports. Then, these topic words are reviewed by sustainability experts, who refine them into well-defined ESG topics. These derived topics are then added to an adapted version of the SASB sustainability taxonomy~\citep{eng2022sasb}.

This approach first involves tokenising each sustainability report into separate sentences, with each sentence transformed into embeddings by Sentence-BERT~\citep{reimers2019sentenceBERT}. BERTopic is then run on these embeddings, employing the KeyBERT configuration. The topic words are then provided to sustainability experts (details of the sustainability experts can be found in~\ref{sec:humanannotate}) who refine the topic words into additional topics. Examples of topic words and their inferred topics can be found in Table~\ref{tab:inference_examples}:

\begin{table}[h]
\centering
\small
\begin{tabularx}{\textwidth}{|X|X|}
\hline
\textbf{Topic keywords} & \textbf{Topic} \\ \hline
strengths, staff, training, opportunities, employees & Workplace Development \\
\hline
cybersecurity, security, cybersecure, authentication & Data Privacy \& Cybersecurity Protection \\
\hline
sustainability, technology, technological, digitalisation & Green Technology \\
\hline
sponsorship, sponsor, charity, scholarship & Community Empowerment \\
\hline
\end{tabularx}
\caption{Examples of Topics Inferred from Topic Keywords}
\label{tab:inference_examples}
\end{table}

Additional topics such as these are then added to an adapted version of the existing SASB framework~\citep{eng2022sasb}, forming a taxonomy of ESG topics (Section~\ref{sec:taxsection}) utilised in ESGSenticNet. 

\subsection{Resulting Taxonomy}\label{sec:taxsection}
\begin{singlespace}
\begin{table}[h]
\centering
\small
\begin{tabularx}{\textwidth}{|X|X|}
\hline
\textbf{Topic Type} & \textbf{Topics} \\ \hline
Pillar & `Environmental', `Social', `Governance'. \\ \hline
Broad & `Development', `Operations', `Workplace', `Outreach', `Management', `Communications', `Compliance' \\ \hline
Sub & `Green Technologies', `Green Infrastructure', `Green Financing', `Sustainable Corporate Operations', `Sustainable Supply Chain', `Sustainable Production Processes', `Sustainable Land Management', `Workplace Wellness', `Workplace Diversity', `Workplace Development', `Community Empowerment', `Strategic Partnerships', `Customer Engagement', `Management Composition', `Management Ethics', `Management Compensation', `Stakeholder Engagement', `Oversight', `Environmental Compliance', `Worker \& Consumer Safety', `Business Compliance', `Data Privacy \& Cybersecurity Protection'\\ \hline
Cross-Broad & `Resource Optimisation', `Waste Management', `Emissions Control', `Ecological Conservation' \\ \hline
Cross-Sub &  `Renewable and Efficient Energy', `Material Sustainability', `Water Conservation', `Wastewater Management', `Solid Waste Management', `Climate Emissions', `Air Quality Emissions'  \\ \hline
\end{tabularx}
\caption{Topic types and topics, with the different topics delineated by a comma}
\label{tab:category_type_descriptions}
\end{table}
\end{singlespace}

We summarise the hierarchical taxonomy derived for ESGSenticNet in Table~\ref{tab:category_type_descriptions}. The taxonomy comprises of topics are divided according to their different topic types--\textit{pillar, broad, cross-broad, sub, cross-sub}, as shown in Table \ref{tab:category_type_descriptions}. 
The taxonomy is structured hierarchically, with topics categorised into different levels based on their scope and relationship. At the highest level, all topics fall under the broad ESG pillars—Environmental (E), Social (S), and Governance (G)—which represent the overarching dimensions of sustainability. Beneath the pillars, topics are grouped into broad and cross-broad topics, which define major sustainability themes. Each broad or cross-broad topic may have constituent sub-topics, which represent more specific areas within that theme. While Table~\ref{tab:category_type_descriptions} summarises our taxonomy, it does not detail the constituent sub-topics of each broad or cross-broad topic. For these relationships as well as topic definitions, please refer to the full descriptions in~\ref{appendix:taxdescr} and Table~\ref{tab:taxonomy}. We elaborate more on the features of this taxonomy below: 

\begin{itemize}
    \item \textbf{Realistic}: Given that we leverage the empirical content from the sustainability reports, our taxonomy comprises additional ESG topics that might otherwise be overlooked by standard frameworks. For instance, `Green Technology' is derived as a taxonomy topic, while not being explicitly emphasised in standard ESG reporting frameworks—i.e. GRI~\citep{machado2021gri}, SASB~\citep{eng2022sasb}. 
    \item \textbf{Activities Cross-Theme}: To capture the interconnectedness of sustainability dimensions~\citep{meuer2020naturecorporatesus}, additional topics include areas of a firm's business activities (`operations', `development' etc.). In contrast with existing frameworks that primarily describe direct sustainability impacts (`emissions control', `waste management' etc.), this offers insight into the sustainability effects of specific business activities. In our taxonomy, topics pertaining to strategic business activities fall under `broad' \& `sub' topic types, and are kept separate from topics pertaining to direct sustainability impacts, which fall under `cross-broad' \& `cross-sub' topic types.
    % \item \textbf{Granularity}: The taxonomy's topics are arranged with a hierarchical structure—i.e. pillars, broad-topics, sub-topics, as explained earlier— allowing analysis at different levels of granularity.
\end{itemize}

\subsection{ESGSenticNet Relations \& Rules}\label{sec:rules}
With the taxonomy established (Table~\ref{tab:category_type_descriptions}), we now define the relation types between concept and pillar/topic (Table~\ref{tab:relation_descriptions}), as well as the rules governing these relations. Each concept within a triple can possess one of the following relations with a pillar/ topic:
\begin{singlespace}
\begin{table}[h]
\centering
\small
\begin{tabularx}{\textwidth}{|X|X|}
\hline
\textbf{Relation} & \textbf{Description} \\ \hline
aligns with & The concept is positively or negatively related to a specific Environmental, Social, or Governance (ESG) pillar. \\ \hline
supports & The concept directly advances the ESG topic with a clear and immediate positive impact. \\ \hline
undermines & The concept directly impedes the ESG topic with a clear and immediate negative impact. \\ \hline
\end{tabularx}
\caption{Relations Types}
\label{tab:relation_descriptions}
\end{table}
\end{singlespace}

The relations, \textit{`supports'} and \textit{`undermines'}, clarify how concepts advance or impede crucial topics of sustainability. In contrast, the relation \textit{`aligns with'} does not consider a concept's impact, highlighting a concept's general connection with broader pillars of environmental, social or governance. At this juncture, it is worthwhile to qualify that the deployment of ESGSenticNet during topic analysis solely uses the `supports' relation (explained further in Section~\ref{sec:experiments}), while the other relations serve as supplementary information. Separately, we now outline the rules that govern all relations in ESGSenticNet:

\begin{itemize}
    \item \textbf{Pillar Assignment:} Each concept `aligns with' a specific pillar, mapping the concepts to the broader sustainability context. 
    \item \textbf{Single Label within Topic Types:} Each concept can hold only one relation with one topic within each topic type.
    \item \textbf{Cross-Labels between Parent \& Children:} Each concept can hold relations with higher-level topics (`broad' and `cross-broad'), as well as with more detailed topics (`sub' and `cross-sub'). 
    \item \textbf{Cross-Labels between Cross \& Non-Cross:} Concepts that `align with' the `environmental' pillar can hold relations with `cross-broad' and `cross-sub' topics, in addition to `broad' and `sub' topics. This reflects the multi-faceted nature of environmental concepts, which can simultaneously impact environmental stewardship (represented by `cross-broad', `cross-sub' topics) and strategic business activities (represented by `broad' and `sub' topics).
\end{itemize}

\subsection{ESGSenticNet Construction through a Neurosymbolic Framework}\label{sec:esgsenticnet_entire_construction}

In sections~\ref{sec:data}~\ref{sec:taxonomy}~\ref{sec:taxsection}~\ref{sec:rules}, we outlined the preliminaries by introducing the data and defining the ESG topics and relations within the knowledge triples (concept, relation, ESG topic) in ESGSenticNet, thereby establishing ESGSenticNet's organisational structure. In this section, we go beyond definitions to detail our neurosymbolic framework for deriving the triples, explaining how concepts are identified and assigned their respective relations.

\begin{figure*}[h]
    \centering
    \includegraphics[width=\textwidth]{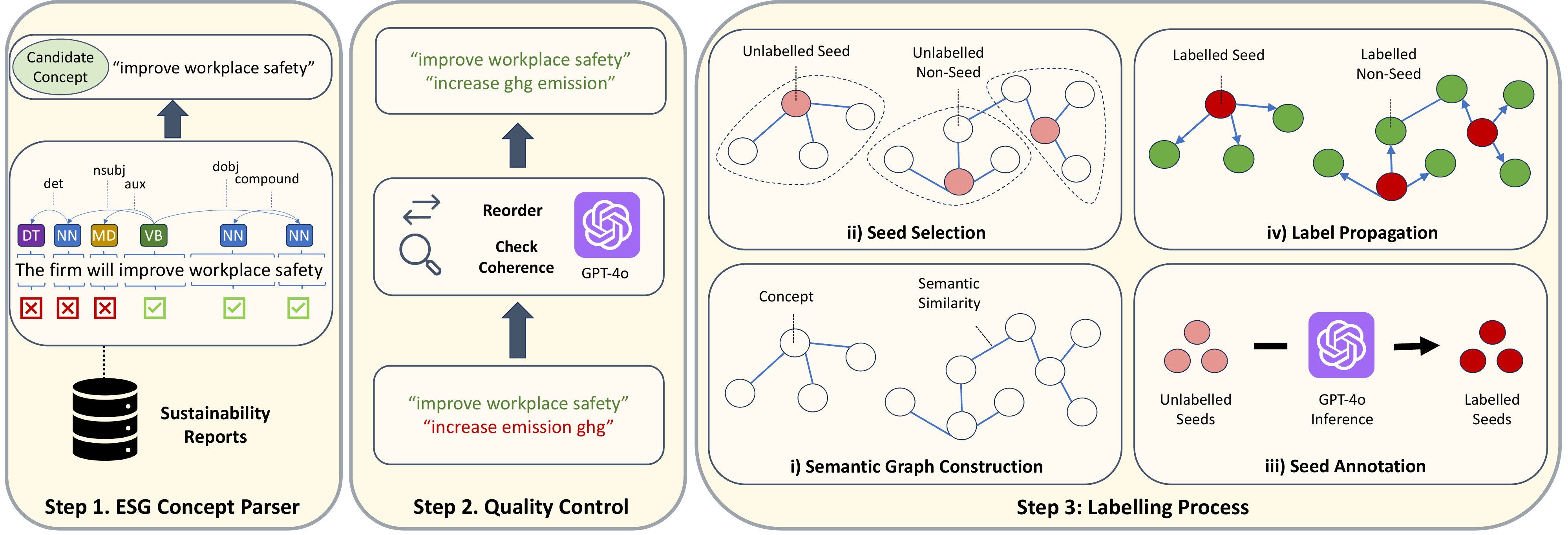}
    \caption{Overview of neurosymbolic framework for ESGSenticNet construction}
    \label{fig:overview}
\end{figure*}

The neurosymbolic framework for constructing ESGSenticNet involves three main phases as shown in figure~\ref{fig:overview}. Step 1: \textit{ESG Concept Parser} (Section~\ref{sect:esgconceptparser}), involves concept parsing, which encompasses extracting candidate concepts from sustainability reports through dependency parsing and part-of-speech tagging. Step 2: \textit{Quality Control} (Section~\ref{sec:processingconcepts}), involves processing coherent concepts from these candidate concepts to ensure their coherence, as a form of quality control. Step 3: \textit{Labelling Process} (Section~\ref{sec:labellingprocess}), involves a labelling methodology to classify the filtered concepts according to the topics within our constructed taxonomy in Section~\ref{sec:taxsection}, leveraging the rules defined in Section~\ref{sec:rules}. 

Accordingly, this labelling methodology in step (3) is detailed in Figure~\ref{fig:overview} (following processes i through iv). These processes integrate GPT-4o inference with a graph-based semi-supervised method, as explained below:
\begin{enumerate}[label=(\roman*)]
    \item \textbf{Semantic Graph Construction} (Section~\ref{sec:semgraphconstruct}): A semantic graph is constructed using the filtered concepts from step (2), leveraging their semantic similarities.  
    \item \textbf{Seed Selection} (Section~\ref{sec:seed select}): Seed concepts are selected from the constructed semantic graph using a seed selection algorithm that applies clustering techniques and graph analysis.  
    \item \textbf{Seed Annotation} (Section~\ref{sec:seed annotation}): Using GPT-4o, the relations between selected seed concepts and taxonomy topics are labelled.  
    \item \textbf{Label Propagation} (Section~\ref{sec:lpa}): The relation labels from seed concepts are propagated to other concepts through a label propagation algorithm.  
\end{enumerate}

Upon completing the labelling process, ESGSenticNet is fully constructed. Notably, this labelling process allows us to label a large amount of diverse concepts despite performing inference on a subset of the concept samples. As a result, computational cost constraints are mitigated, as we are able to label concepts more extensively. This, in turn, significantly increases the extensiveness and diversity of our knowledge base, enhancing its effectiveness as a lexical method for ESG topic analysis.

% (sections~\ref{sec:seed select},~\ref{sec:seed annotation},~\ref{sec:semgraphconstruct},~\ref{sec:lpa})

\subsection{ESG Concept Parser}\label{sect:esgconceptparser} 
Multiword expressions are linguistic phenomena that refer to common linguistic constructions such as idioms, collocations and sayings~\citep{constant2017multiword}. In the context of sustainability analysis, multiword expressions frequently communicate key sustainability ideas~\citep{smeuninx2020susreportstedious}.  We refer to these multiword expressions as ESG ``concepts''. Extracting these concepts or ``concept extraction'' has been explored for metaphors~\citep{mao2023metaproonline}, finance~\citep{du2023finsenticnet} and sentiment analysis~\citep{cambria2022sentic}. Yet, to the best of our knowledge, concept extraction within the ESG domain has not been studied. Therefore, to capture ESG concepts, a new concept parser is required, with an explicit focus on the distinct elements of sustainability discourse. These include the following designs:
\begin{itemize}
    \item \textbf{Complex Nominals, Modifiers, Nouns \& Adjectives}: Sustainability discourse is complex, containing a high concentration of adjectives and nouns which commonly function as modifiers within nominal phrases~\citep{smeuninx2020susreportstedious}. Phrases such as `renewable energy', `carbon emissions', demonstrate constructions that efficiently convey complex sustainability ideas.
    \item \textbf{Descriptive \& Prescriptive Elements}: Sustainability discourse often involves descriptive and prescriptive language. The former (i.e. stative verbs like `invested', `reduced') is utilised when a company reports its current sustainability activities~\citep{papoutsi2020susdisclosuresdescriptive}, while the latter (i.e. verbs of volition like `plan', `commit') is observed when a company pledges future sustainability activities~\citep{silva2021corporateprescriptive}. 
\end{itemize}

To effectively extract these unique aspects of sustainability language, we formalise ESG concepts through the following syntactic structure--verbs acting on complex noun phrases, where the words involved function as modifiers, descriptive and prescriptive elements. Specifically, the concepts are two or three word phrases, typically beginning with a $verb$, followed by $noun + noun$, or $adjective + noun$ combinations with specific grammatical dependencies. These dependencies include: $verb-noun$ connection via \textit{(nsubj, obj, obl)}, $adjective-noun$ or $noun-noun$ connection via \textit{(compound, amod, nn, appos, flat, nmod)}. This framework underpins the ESG Concept Parser algorithm (\ref{algo:esgconceptparser}), designed to capture key ideas within sustainability reports, particularly ESG efforts. For instance, \textit{`improve (VB) workplace (NN) safety (NN)'}, where \textit{`workplace'} is the $compound$ of \textit{`safety'} and \textit{`safety'} is the $dobj$ of \textit{`improve'}. 

\begin{singlespace}
\begin{algorithm}[H]
\caption{ESG Concept Parser}
\small
\label{algo:esgconceptparser}
\begin{algorithmic}[1]
    \State \textbf{Input:} Sentence from sustainability report
    \State \textbf{Output:} ESG Concepts
    \State POS tag the sentence
    \State Dependency parse the sentence
    \For{\textit{dependency, head, dependent} in sentence}
        \If{dependency $\in \{\textit{nsubj}, \textit{obj}, \textit{obl}\}$}
        \If{\textit{head} is \underline{\textit{noun}} \& \textit{dependent} is \underline{\textit{verb}}}
        \State add to Phrases: \textit{dependent+head} as \textit{\underline{verb}+\underline{noun}} combination
        \EndIf
        \EndIf
        \EndFor
    \For{\textit{$word_{1}$, $word_{2}$ in Phrases}} 
        \For{\textit{dependency, head, dependent} in sentence}
            \If{\text{head} is \underline{\textit{noun}} \& dependency $\in \{\textit{compound}, \textit{amod}, \textit{nn}, \textit{appos}, \textit{flat}, \textit{nmod}\}$}
            \If{\textit{dependent} is \underline{\textit{noun}}}
            \State add to Phrases: \textit{$word_{1}$}\textit{+dependent+}\textit{$word_{2}$} as \textit{\underline{verb}+\underline{noun}+\underline{noun}} combination 
            \EndIf
            \If{\textit{dependent} is \underline{\textit{adj}}}
            \State add to Phrases: {$word_{1}$}\textit{+dependent+}\textit{$word_{2}$} as \textit{\underline{verb}+\underline{adj}+\underline{noun}} combination
            \EndIf
            \EndIf
        \EndFor
    \EndFor
\end{algorithmic}
\end{algorithm}
\end{singlespace}

\subsection{Quality Control}\label{sec:processingconcepts}
To minimise computational cost, parsed concepts from algorithm (1) are filtered to retain the top 110k most frequently occurring ones. Of these, we eventually annotate 23k concepts based on our estimated computational resources, which dictates the number of seeds we select within our semantic graph. Prior to the annotation process, these concepts must undergo processing to ensure their coherence, as a form of quality control. This involves utilising GPT-4o in few-shot setting (Brown, 2020). GPT-4o is leveraged due to the capacity of large generative LLMs to infer~\citep{wei2022emergent}, as well as their powerful natural language understanding capabilities~\citep{yang2024harnessing}. Using GPT-4o to process concepts for coherence involves conducting the following inference tasks in order: (1) reorder words, if required, to improve the coherence of the concept, (2) determine if the subsequent concept is intelligible. We provide the prompt for (1) and (2) through Figure~\ref{prompt:reorder} and Figure~\ref{prompt:coherence} respectively.

\begin{singlespace}
\begin{figure}[H]
    \centering
    \small
    \begin{tcolorbox}[colframe=black, colback=white, boxrule=0.8pt, width=\textwidth]
    \textbf{[Task Description]} \\
    As a language expert, your task is to rearrange words in phrases to enhance their coherence and intelligibility if necessary. \\[5pt]
    
    \textbf{[Response Instructions]} \\
    Please list all your outputs for all the phrases given. Follow these steps for each given phrase: Assess whether reordering the words in the phrase can improve its coherence and intelligibility. If needed, reorder the words to achieve the best clarity and coherence, but do not add new words and do not change the words. If no reordering is needed, retain the original order. Record the output accordingly. \\[5pt]
    
    \textbf{[Response Format]} \\
    Fill in your response in the triple backticks below using the following format: \\
    \texttt{Input: <original phrase>} \\
    \texttt{Output: <reordered or original phrase>}
    \end{tcolorbox}
    \caption{Prompt for reordering an ESG concept for coherence}
    \label{prompt:reorder}
\end{figure}

\begin{figure}[H]
    \centering
    \small
    \begin{tcolorbox}[colframe=black, colback=white, boxrule=0.8pt, width=\textwidth]
    \textbf{[Task Description]} \\
    As a language expert, you are tasked with determining if a phrase is intelligible. \\[5pt]
    
    \textbf{[Response Instructions]} \\
    Please list all your outputs for all the phrases given. Follow these steps for each given phrase: Assess if a given phrase is intelligible. If the phrase is intelligible, output a `True' for the phrase. If the phrase is not intelligible, output a `False' for the phrase. \\[5pt]
    
    \textbf{[Response Format]} \\
    Fill in your response in the triple backticks below using the following format: \\
    \texttt{Input: <phrase>} \\
    \texttt{Output: <True/False>}
    \end{tcolorbox}
    \caption{Prompt for evaluating an ESG concept's coherence}
    \label{prompt:coherence}
\end{figure}
\end{singlespace}

\subsection{Labelling Process}\label{sec:labellingprocess}

\subsubsection{Semantic Graph Construction}\label{sec:semgraphconstruct}

\begin{equation}
    G = (V, E)
    \label{eqn:graphini}
\end{equation}
\begin{equation}
    V = \{\mathbf{e}_i \mid \mathbf{e}_i = \text{S-BERT}(cp_i)\}
\end{equation}
\begin{equation}
    \text{sim}(\mathbf{e}_i, \mathbf{e}_j) = \frac{\mathbf{e}_i \cdot \mathbf{e}_j}{\|\mathbf{e}_i\| \|\mathbf{e}_j\|}
\end{equation}
\begin{equation}
    E = \{ (\mathbf{e}_i, \mathbf{e}_j) \mid \text{sim}(\mathbf{e}_i, \mathbf{e}_j) > 0.80 \}
    \label{eqn:graphfin}
\end{equation}
\begin{equation}
    w_{ij} = \text{sim}(\mathbf{e}_i, \mathbf{e}_j), \quad \forall (\mathbf{e}_i, \mathbf{e}_j) \in E
    \label{eqn:edgeweight}
\end{equation}

Processed concepts undergo labelling to derive their relationships with different ESG topics. Given the exceedingly large number of concepts, annotating all concepts would be prohibitively expensive. To overcome this, we devise a semi-supervised framework that enables a large volume of diverse concepts to be labelled from annotating only a subset of all concepts. Via a semantic graph, we leverage the semantic similarities between concepts to propagate labels from annotated concepts (seeds) to unlabelled concepts (non-seeds). This graph is constructed by transforming each concept $cp_i$ into embeddings $e_i$ through S-BERT~\citep{reimers2019sentenceBERT}, with edges constructed by leveraging a cosine similarity threshold of 0.80 between each node $e_i$. To qualify the strength of relation between nodes (concepts), the edge weights $w_{ij}$ are defined as the cosine similarity scores between the nodes. We describe the full graph via equations~(\ref{eqn:graphini}) through~(\ref{eqn:edgeweight}).

\subsubsection{Seed Selection}\label{sec:seed select}
Within our semantic graph, seeds are selected for annotation through data clustering and graph analysis. Given the semi-supervised label propagation from seeds to non-seeds, we formulate the following objectives for seed selection: \noindent \textbf{(I)} Seeds must be semantically diverse to ensure that diverse labelled samples are yielded post label propagation. This enhances the variety and comprehensiveness of labelled concepts in our knowledge base. \noindent \textbf{(II)} Seeds must propagate their labels to as many non-seeds as possible. This maximises the number of labelled samples we can attain while minimising annotation costs. To achieve (I) and (II), we utilise \textbf{Data Clustering} and \textbf{Graph Analysis}, with these components culminating in a \textbf{Seed Selection Algorithm}. 

\begin{equation}
    CQI = \frac{|\{e \in V \mid \alpha_e > 0.60\}|}{|V|}
    \label{eqn:cqi}
\end{equation}

\noindent \textbf{Data Clustering.} To accomplish (I), the embeddings of concepts $e_i$ are clustered to enable seed selection from diverse semantic spaces. This clustering involves projecting the embeddings to a lower dimensional space via UMAP~\citep{mcinnes2018umap}, before running HDBSCAN~\citep{campello2013hdbscan} on these projected embeddings. While HDBSCAN is flexible to the shape and number of clusters, and robust to noise~\citep{campello2013hdbscan}, reducing the dimensionality of embeddings ensures good clustering performance. This is because HDBSCAN relies on distance-based metrics which lack meaningfulness in the high dimensional spaces of textual embeddings~\citep{aggarwal2001surprisingdistance}. The embeddings are projected to lower dimensions for the sole purpose of clustering. In our later section, label propagation leverages the default dimensions of S-BERT embeddings. Clusters are evaluated via a Confidence Quality Index (\( CQI \)), which denotes the proportion of embeddings, $e$, out of all embeddings, $V$, with a clustering confidence level $\alpha_e$ beyond 0.60, as described by equation~(\ref{eqn:cqi}). For our use of HDBSCAN, this simple metric is more appropriate than others like Slihouette Score~\citep{shahapure2020silhouette}, which assumes that all embeddings belong to a cluster. Hyperparameter tuning is done via random search, with the best run having a $CQI$ of 0.66, UMAP parameters $n$\_$neighbours$$=$32, $n$\_$components$$=$2, and the HDBSCAN parameter $min$\_$cluster$\_$size$$=$2. 

\begin{equation}
    \begin{split}
        Q(\mathbf{e}_i) = | \left\{ (\mathbf{e}_i, \mathbf{e}_j) \in E \mid (\mathbf{e}_i \in \mathbf{c_k})  \right. \\
        \left. \wedge \left(\forall \mathbf{e}_{s} \in S_{c_k}, (\mathbf{e}_s, \mathbf{e}_j) \notin E\right) \right\} |
    \end{split}
    \label{eqn:conscore}
\end{equation}

\noindent \textbf{Graph Analysis.} To accomplish (II), graph analysis is utilised to select seeds that are the most consequential nodes. Given that edges facilitate label propagation, a node is considered more consequential if it possesses a greater number of unique edges. Unique edges are defined as connections to nodes that do not share an edge with the already selected seed nodes within the same cluster. We formulate the local consequential score $Q(e_i)$, where $S_{c_k}$ is the set of nodes already selected as seeds within the same cluster, $c_{k}$, as the node $e_i$, with this described in equation~(\ref{eqn:conscore}). \\

\noindent \textbf{Seed Selection Algorithm.} Capitalising on \textbf{Data Clustering} and \textbf{Graph Analysis}, an algorithm (\ref{algo:seed_selection}) is developed for seed selection to achieve objectives (I) \& (II). Given a target number of seeds to label, $T$, and the embeddings of all ESG concepts, $V$, seeds, $S_{total}$, are selected from clusters, $C$. Each cluster, $c_k$, within $C$, is a set of embeddings of concepts, $e_i$.  Within each $c_k$, seeds of a cluster, $S_{c_k}$, are selected by taking the embedding with the maximum local consequential score $Q(e_i)$ at each selection step. This maximises the selection of seeds with unique edges to increase label propagation to non-seeds. Additionally, each cluster should ideally possess an equivalent proportion, $P$, of seeds, and at least one seed each. We accomplish this by iteratively adjusting $P$ according to parameter $\beta$ (ESGSenticNet uses $\beta = 0.01$), to ensure that our estimated number of seeds $N_{s}$ is as close to our target number as possible $T$ (the difference between $N_{s}$ and $T$ denoted by $\Delta$). From $P$, the specific number of seeds for each cluster, $n_{c_k}$ can be computed.  This ensures that seeds are fairly represented across different clusters to enable the semantic diversity of seeds. 

\begin{singlespace}
\begin{algorithm}[H]
\small
\caption{Seed Selection}
\label{algo:seed_selection}
\begin{algorithmic}[1] % The number [1] indicates line numbers are shown
\State \textbf{Input:} $V$, $T$, $C$
\State \textbf{Output:} $S_{\text{total}}$
    \State $N \gets$ $|V|$
    \State $P \gets \frac{T}{N}$
    \State $\Delta_{prev} \gets \infty$
    \While{True}
        \State $N_{s} \gets \sum_{c \in C} \max(1, \lfloor P \cdot |c| \rfloor)$ 
        \State $\Delta_{curr} \gets |N_{s} - T|$
        \If{$\Delta_{curr} \geq \Delta_{prev}$}
            \State \textbf{break}
        \Else
            \State $\Delta_{prev} \gets \Delta_{curr}$
        \EndIf

        \If{$N_{s} < T$}
            \State $P \gets P + \beta$
        \Else
            \State $P \gets P - \beta$
        \EndIf
    \EndWhile
    \For{$c_k \in C$}
        \State $n_{c_k} \gets \max(1, \lfloor P \cdot |c_k| \rfloor)$
        \State $S_{c_k} \gets \emptyset$
        \While{$|S_{c_k}| < n_{c_i}$}
            \State $e_{\max} \gets \arg\max_{e \in c_k} Q(e)$   
            \State $c_k \gets c_k \setminus \{e_{\max}\}$ 
            \State $S_{c_k} \gets S_{c_k} \cup \{e_{\max}\}$
        \EndWhile
        \State $S_{total} \gets S_{total} \cup S_{c_k}$ 
    \EndFor
    \State \Return $S_{\text{total}}$ 
\end{algorithmic}
\end{algorithm}
\end{singlespace}

\subsubsection{Seed Annotation}\label{sec:seed annotation}

\begin{figure}[H]
    \centering
    \includegraphics[width=0.6\textwidth]{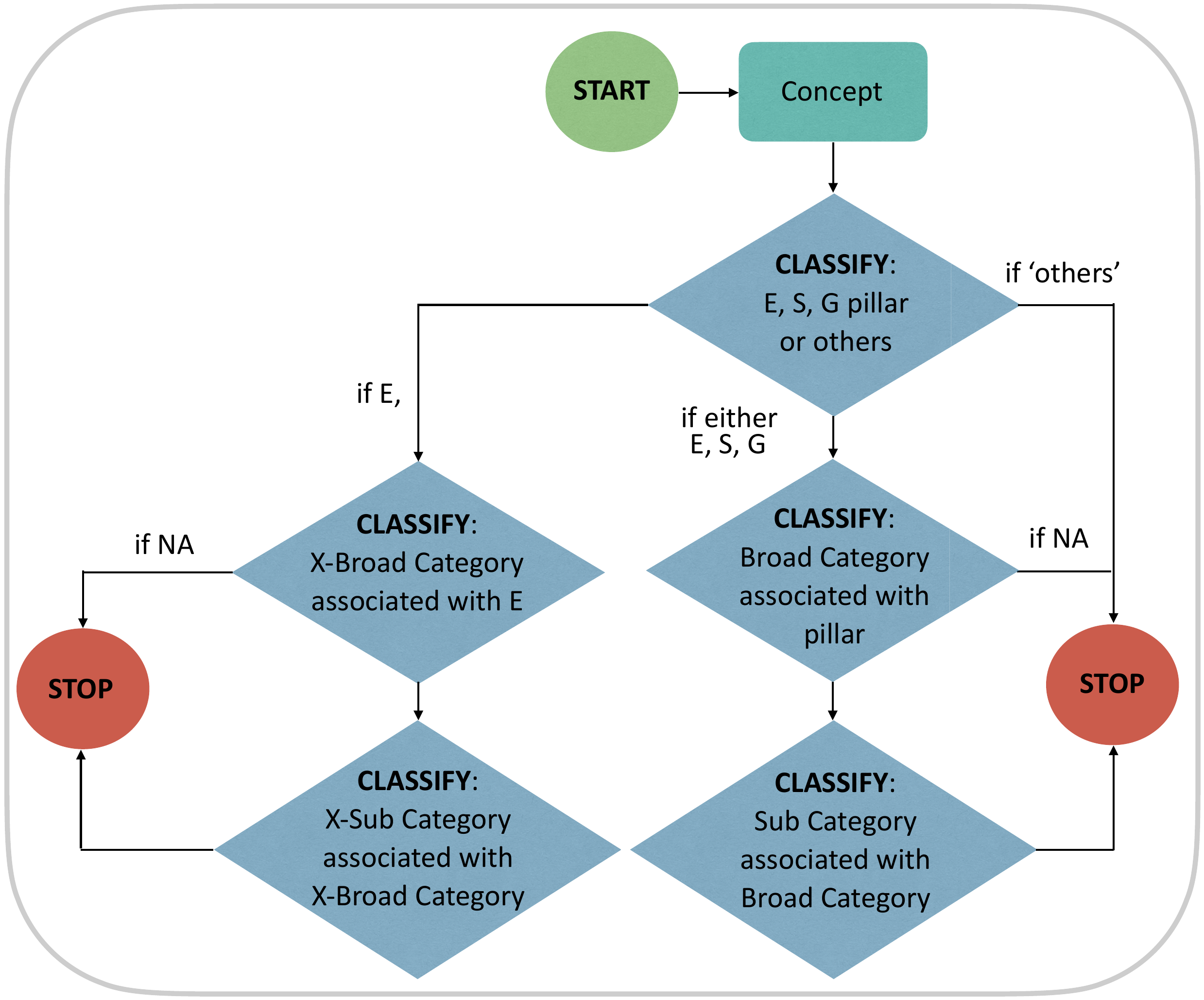}
    \caption{Flowchart depicting inference process}
    \label{fig:inference_flowchart}
\end{figure}

Through GPT-4o few-shot prompting, the selected seeds are annotated with their relations (`aligns with', `supports', `undermines') with respect to a pillar or topic, before their labels are later propagated within the semantic graph (Section~\ref{sec:lpa}). ESGSenticNet retains a considerable number of seed concepts ($>16k$) directly labelled through GPT-4o. This improves the accuracy of ESGSenticNet's relations, given that the reasoning capabilities of GPT-4o~\citep{huang2022llmreasoning} allows it to infer relations for concepts outside of similar semantic spaces. For example, consider the concepts `use led light', `achieve energy efficiency' categorised under `Resource Optimisation'. Textual similarity methods leveraging these for comparison might not categorise `emphasise water importance' similarly, despite its relevance. However, using GPT-4o, `emphasise water importance' is appropriately labelled, due to the LLM's capacity to infer semantic relationships beyond mere textual similarities. 

% ESGSenticNet's approach in the labelling process (step (3) in Figure~\ref{fig:overview}) focuses on seed annotation before leveraging textual similarities to expand the initial seed corpus. 
% Similar to popular sustainability lexicons~\citep{baier2020dict, tian2023naturesusdataset}, 
% Moreover, GPT4o as an automated technique allows for a large number of concepts to be labelled, and this labelling is then considerably increased by our graph-based semi-supervised method, all within step (3) in Figure~\ref{fig:overview}. overcoming prohibitive human labelling costs. This allows us to derive an extensive knowledge base.

To improve GPT-4o classification accuracy, inference tasks are divided into smaller, more manageable sub-tasks~\citep{chang2024llmperformancesurvey}. These tasks align with the taxonomy’s hierarchy levels, following the flowchart in Figure~\ref{fig:inference_flowchart}. Subsequent categorisations of seeds are contingent upon their initial classification into broader topics, and it aligns with the predefined associations between the ESG pillars and broad topics, as well as the connections between the broad and sub topics. Prompts for these tasks are highlighted in Figure~\ref{prompt:domain} (pillar labelling), Figure~\ref{prompt:relation} (relation-topic labelling).

\begin{singlespace}
\begin{figure}[H]
    \centering
    \begin{tcolorbox}[colframe=black, colback=white, boxrule=0.8pt, width=\textwidth]\small
    \textbf{[Task Description]} \\
    As a sustainability expert, your task is to classify phrases into the topics of Environmental, Social, Governance, or Others based on their explicit relevance to the Environmental, Social, Governance topics. \\[5pt]
    
    \textbf{[Definitions of topics]} \\
    \textit{Definitions of ESG according to taxonomy, with the inclusion of `Others'} \\
    \textbf{Others}: Reserved for phrases that do not meet the explicitness and specificity requirements for Environmental, Social, or Governance. This includes phrases that are vague, ambiguous, or cover multiple Environmental, Social, Governance topics without clear categorisation. \\[5pt]
    
    \textbf{[Response Instructions]} \\
    Evaluate each phrase's explicit content to determine its most appropriate topic. Assign each phrase to only one topic: Environmental, Social, Governance, or Others. Only phrases that are clear and explicit can be categorised under Environmental, Social, or Governance. Phrases that are vague or not clearly related to these topics should be categorised as Others. Ensure all responses strictly adhere to the definitions and format provided. \\[5pt]
    
    \textbf{[Response Format]} \\
    Fill in your response in the triple backticks below using the following format: \\
    \texttt{Input: <phrase>} \\
    \texttt{Response: <Environmental, Social, Governance, or Others>}
    \end{tcolorbox}
    \caption{Prompt for labelling an ESG concept's pillar}
    \label{prompt:domain}
\end{figure}

\begin{figure}[H]
    \centering
    \small
    \begin{tcolorbox}[colframe=black, colback=white, boxrule=0.8pt, width=\textwidth]
    \textbf{[Task Description]} \\
    As a sustainability expert, you are tasked with evaluating the relationship between phrases and ESG-related topics. \\[5pt]

    \textbf{[Definitions of Relationships]} \\
    \textbf{supports}: The phrase, specific and explicit to the topic, directly advances it with a clear and immediate positive impact. \\
    \textbf{undermines}: The phrase, specific and explicit to the topic, directly impedes its goals with a clear and immediate negative impact. \\[5pt]

    \textbf{[Definitions of Topics]} \\
    \textit{Definitions of topics according to taxonomy} \\[5pt]

    \textbf{[Response Instructions]} \\
    Evaluate the explicit content of each phrase to determine the most appropriate and accurate relationship and topic. Identify the single most relevant and appropriate relationship and topic. If no specific and explicit relationship can be determined between the phrase and any topic, or if the phrase does not fit any topic due to vagueness or irrelevance, classify the phrase as `not applicable'. \\[5pt]

    \textbf{[Response Format]} \\
    For each phrase, provide your analysis using the format below. Output only one tuple if applicable, otherwise state `not applicable'. \\
    \texttt{Input: <phrase>} \\
    \texttt{Response: <(relationship, topic)> or `not applicable'}
    \end{tcolorbox}
    \caption{Prompt for labeling an ESG concept's relation with a topic}
    \label{prompt:relation}
\end{figure}
\end{singlespace}

% For instance, if a esg concept initially `aligns with' the `social' pillar, only `social'-related broad categories , such as (`Workplace', `Outreach', `Communications'), are included in the subsequent inference prompt. Similarly, if a phrase `supports' or `undermines' the broad category of `Outreach', the next inference prompt will only contain its children sub categories such as (`Community Empowerment', `Strategic Partnerships', `Customer Engagement'). Notably, a phrase that `aligns with' `environmental', will undergo an additional inference pipeline involving cross-broad (X-Broad) categories and children cross-sub (X-Sub) categories. This is because we allow cross-labelling for phrases aligned with `environmental', as explained earlier (see Section~\ref{sec:relationcat}). 

\subsubsection{Label Propagation}\label{sec:lpa}

\begin{equation}
L^{(k+1)} = D^{-1/2} A D^{-1/2} L^{(k)},
\end{equation}
\begin{equation}
l_i^{(k+1)} = l_i^{(0)} \quad \forall i \leq m
\end{equation}

Annotated seeds have their labels propagated to non-seeds via the label propagation algorithm~\citep{wang2021lpa}. This increases the labelled data within our knowledge base without incurring additional labelling costs. We utilise the label propagation algorithm for its ability to perform soft transducive inference — iteratively proliferating labels across the graph probabilistically, based on global similarity~\citep{wang2021lpa}. This contrasts with other algorithms — i.e. k-nearest neighbours~\citep{peterson2009knn}, which assign hard labels without modeling uncertainty (i.e. via probability). In our context, the label propagation algorithm's incorporation of probability allows it to better capture semantic nuances and ambiguity across concepts. This is useful when one concept may relate to multiple categories. Specifically, the following algorithm is iterated until convergence, where $A$ is an adjacency matrix containing $a_{i,j}$ that denotes the edge weight between connected nodes $e_{i}$ and $e_{j}$. Each node represents an embedding of a concept, while the edge weight denotes cosine similarity between concepts. $L$ is a matrix that represents the relations labels of each node with respect to a topic (i.e. \textit{supports} within the (`reduce water consumption', \textit{supports}, `resource optimisation'). $m$ is the number of nodes possessing labels and $D$ is a diagonal matrix containing $d_{i, i}$, which equals the sum of elements in row $i$ of $A$. We use parameters $n\_layers$=50, $\alpha$=0.5.

% On top of knowledge triples, ESGSenticNet appends polarity labels to concepts. This enables potential use in sentiment analysis. 

% \subsection{Expansion via WordNet \& Wordsense}

% \subsection{Further additions to ESGSenticNet}~\label{sec:polarityaddition}
% To increase the generalisability of ESGSenticNet for more tasks, including sentiment analysis, ESGSenticNet also appends polarity labels to concepts, on top of the knowledge triples. For triples already comprising of the `supports' and `undermines' relations, concepts within these triples can directly inherit the `positive' and `negative' labels. Concepts without these relations are put through GPT-4o inference for polarity classification, using the prompt in figure~\ref{prompt:polarity}. Human evaluation demonstrates that the accuracy of the polarity labels of a 500 concept sample in ESGSenticNet is 90.6\%. 

\subsection{ESGSenticNet — Statistics \& Samples}

\begin{figure*}[h]
    \centering
    \includegraphics[width=0.9\textwidth]{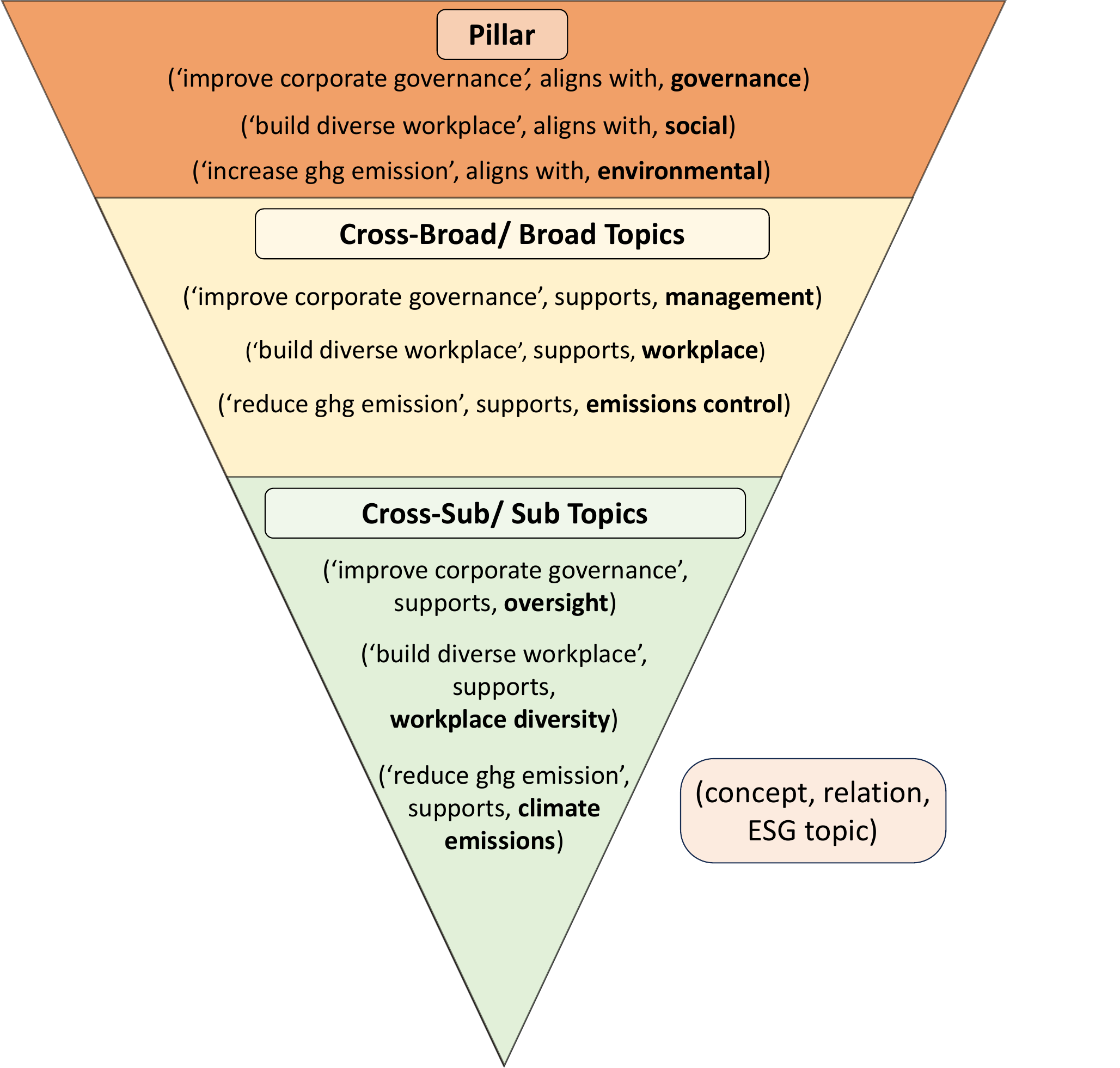}
    \caption{Overview of ESGSenticNet}
    \label{fig:ESGSenticNet_example}
\end{figure*}

After label propagation, ESGSenticNet is fully constructed. We provide an visualisation of ESGSenticNet in Figure~\ref{fig:ESGSenticNet_example}. ESGSenticNet comprises 44232 triples containing 23245 unique ESG concepts. Of the 23245 concepts, 16011 are seeds and 7234 are non-seeds. We provide samples of ESGSenticNet's knowledge triples (\textit{concept}, \textit{relation}, \textit{ESG topic}) in Table~\ref{tab:ESGSenticNet Samples}.

\begin{singlespace}
\begin{table}[H]
\centering
\small
\begin{tabularx}{\textwidth}{|X|X|X|}
\hline
\textbf{Concept} & \textbf{Relation} & \textbf{ESG topic} \\ \hline
build diverse workplace & supports & workplace diversity\\
\hline
halve carbon emission & supports & emissions control \\
\hline
involve workplace injury & undermines & worker \& consumer safety compliance \\
\hline
minimise resource consumption & supports & resource optimisation \\
\hline
organise charity event & supports & outreach \\
\hline
produce chemical waste & undermines & operations\\
\hline
\end{tabularx}
\caption{Samples of knowledge triples from ESGSenticNet}
\label{tab:ESGSenticNet Samples}
\end{table}
\end{singlespace}

Table~\ref{tab:esgsenticnet_meta_stats} provides a meta statistic of of how ESGSenticNet compares with existing sustainability dictionaries—i.e Baier~\citep{baier2020dict}, Naiara~\citep{Naiara2024news}, Kang~\citep{kang2022finsim4terms}. To qualify, the concepts within ESGSenticNet's triples are taken as ``terms''.

\begin{singlespace}
    \begin{table}[h]
    \centering
    \small
    \begin{tabular}{|l|l|}
    \hline
    \textbf{Sustainability Dictionaries/ Knowledge Base(s)} & \textbf{Total Unique Terms} \\ \hline
    ESGSenticNet & \textbf{23245} (within 44232 triples) \\ \hline
    Baier & 491 \\ \hline
    Naiara & 167 \\ \hline
    Kang* & 792 \\ \hline
    \end{tabular}
    \caption{Total number of terms of different sustainability corpora, *indicates not publicly available.}
    \label{tab:esgsenticnet_meta_stats}
    \end{table}
\end{singlespace}

Finally, we provide a full statistical breakdown of ESGSenticNet in Appendix~\ref{appendix:c} Table~\ref{tab:stats}. This outlines the total triples corresponding to each ESG topic, with a breakdown of the relations involved.

\subsection{Ablation Studies}\label{sec:ablation}

\subsubsection{Effectiveness of ESGSenticNet Seed Selection Algorithm}

In this baseline study, we assess the extent of label propagation through ESGSenticNet's seed selection algorithm, against naive methods like selecting seeds randomly or based on highest degree centrality scores. Each method is run on the constructed semantic graph (Section~\ref{sec:semgraphconstruct}), to select different batches of 10k seeds. Aligning with majority of the classification tasks in the main study, three different psuedo-labels are used, and randomly assigned to the selected seeds. Label propagation is run to observe the number of newly propagated labels. Although propagation would be more accurate with real labels, random pseudo-labelling provides an unbiased estimate of propagation extensiveness irrespective of label content, and reduces the cost of extensive testing. Moreover, the entire process is repeated over 100 trials, showing that across each method, the number of newly propagated labels remain consistent despite the randomness of seed labelling.

\begin{singlespace}
\begin{table}[h]
\centering
\small
\begin{tabular}{|l|l|}
\hline
\textbf{Seed Selection Method} & \textbf{Propagated Labels} \\ \hline
Highest Degree Centrality & 4686  \\ \hline
Random Selection & 7456 \\ \hline
ESGSenticNet & \textbf{9617}\\ \hline
\end{tabular}
\caption{Effectiveness of Label Propagation, based on the number of newly propagated labels}
\label{tab:esgsenticnet_labeleff}
\end{table}
\end{singlespace}

Compared to naive methods, from Table~\ref{tab:esgsenticnet_labeleff}, we observe that ESGSenticNet's seed selection algorithm results in a greater number of newly propagated labels. 

\subsubsection{GPT-4 Human Expert Agreement on Topic Analysis Evaluation Metrics}~\label{sec:agreement}
GPT-4 evaluations have shown significant reliability in the literature~\citep{zheng2023llmjudging}. In line with established research practices~\citep{rafailov2024direct}, we validate the GPT-4 evaluations in our use case by conducting a GPT4-human agreement study. Specifically, a study is conducted to assess GPT-4's ability for the following tasks-- i) classifying topic terms as ESG-related, ii) classifying topic terms as actions taken toward ESG. A random sample of 510 topic terms were gathered for each task. GPT-4 annotations for each set of topic terms are compared with annotations from 2 different groups of 3 human experts each, with each group handling a different task. These human experts comprise of doctoral candidates actively conducting research in sustainability finance, and thus possess domain expertise well beyond the level required for this validation task. Moreover, to ensure the robustness of their judgements, the human experts were deliberately kept independent from the paper's development. An average agreement score of 83.7\% observed for task i) and 91.8\% for task ii). While this highlights a reasonable level of robustness for GPT-4 evaluations in our study, it also underscores the broader potential of GPT-4 as a scalable and reproducible evaluation tool in the ESG domain.

For evaluating GPT-4 human agreement, the following instructions were given for classifying topic terms as ESG-related.

\begin{singlespace}
\noindent\fbox{%
  \begin{minipage}{\dimexpr\columnwidth-2\fboxsep-2\fboxrule}
    \small The table below presents various terms. Determine if each term is ESG-related. Mark `TRUE' if the term is ESG-related and `FALSE' otherwise.
  \end{minipage}
}
\end{singlespace}

For evaluating GPT-4 human agreement, the following instructions were given for classifying topic terms as an action taken toward ESG.

\begin{singlespace}
\noindent\fbox{%
  \begin{minipage}{\dimexpr\columnwidth-2\fboxsep-2\fboxrule}
    \small The table below presents various terms. Determine if each term expresses an action toward improving a company's ESG performance. Mark `TRUE' if the term express an action toward improving ESG performance and `FALSE' otherwise.
  \end{minipage}
}
\end{singlespace}

\subsubsection{Effectiveness of ESGSenticNet Flexible Matching Method}\label{sec:flexible}
A study is conducted to assess the precision of matching ESGSenticNet concepts through the `flexible' method depicted in Section~\ref{sec:experiments}. A random sample of 200 sentences matched through this method was collected, excluding sentences matched via the `precise' method. Human annotators were tasked with evaluating whether each sentence clearly conveys the idea expressed represented by its matched concept. From our results, 75.0\% of sentences comprised the idea expressed by its corresponding matched concepts. This suggests that the `flexible' method retains a moderately high level of precision, despite allowing for greater semantic variability. For evaluating the ESGSenticNet flexible matching method, the following instructions were given to evaluate if each sentence clearly conveys the idea expressed by its matched concept.
\begin{singlespace}
\noindent\fbox{%
  \begin{minipage}{\dimexpr\columnwidth-2\fboxsep-2\fboxrule}
  \small The table below presents different sentences and a corresponding concept. Determine if each sentence clearly conveys the idea expressed by its matched concept. Mark `TRUE' if the sentence clearly conveys the concept's idea, and `FALSE' otherwise.
  \end{minipage}
}
\end{singlespace}

\section{Literature Review}\label{sec:litreview}

\subsection{NLP-driven ESG topic analysis.}
To extract insights from sustainability reports, NLP tools can be deployed for different tasks, such as ESG topic analysis~\citep{ferjanvcivc2024textualbertopicesg,li2022sustainabilityfashion}, sentiment classification~\citep{pasch2022sentimentclass}, extracting textual patterns~\citep{bi2023lexical}, with ESG topic analysis being a key task~\citep{ong2024xnlpsusanalysis}. ESG topic analysis facilitates ESG assessments in several ways. Specifically, analysing topics allows stakeholders to monitor corporate alignment with UN Sustainability Development Goals~\citep{WANG2020sdg}, to analyse the future orientation of sustainability reports~\citep{HEICHL2023textmining}, and to construct metrics with respect to ESG performance~\citep{tian2023naturesusdataset}. 

Current NLP methods for this task predominantly include rule-based lexical methods that leverage existing sustainability dictionaries such as Baier~\citep{ignatov2023esgdict}, and unsupervised topic modeling that can be adapted to ESG analysis~\citep{zhou2021shipping, MANDAS2023esgtopicmodel}. Yet, while having advantages as automated tools~\citep{abram2020methods}, these methods may not be tailored for the specialised requirements of ESG topic analysis. Moreover, building specific tools for ESG topic analysis has been under-explored, with efforts primarily focusing around the creation of sustainability lexicons~\citep{baier2020dict, Naiara2024news, kang2022finsim4terms}. However, these lexicons have notable limitations, which we will discuss further in the next section.

\subsection{Lexicons for ESG Analysis}\label{sec:litrev_lexicon}
% Knowledge base construction within sustainability
% like  SenticNet~\citep{cambria2010senticnet} or ConceptNet~\citep{liu2004conceptnet}
Lexical approaches have been widely used to analyse sustainability reports~\citep{HEICHL2023textmining}, given their effectiveness in corporate text analysis and accessibility to stakeholders with diverse technical backgrounds~\citep{nguyen2022textualcoporatebankruptcy,bochkay2023textualaccounting}. To facilitate lexical analysis, lexicons specific to ESG analysis have been developed. For instance, approaches such as Baier's dictionary~\citep{baier2020dict}, have focused on developing single-word lexicons according to ESG pillars. However, while Baier's dictionary has been a popular choice for ESG analysis studies~\citep{ignatov2023esgdict}, its single-word lexicons cannot adequately capture the semantics of the sustainability field~\citep{smeuninx2020susreportstedious}. In contrast, dictionaries that have collected ESG-related text spans (of multiple-grams)—i.e. ~\citet{du2024esgbigramtext},~\citet{kang2022finsim4terms}—possess limited ESG topics or are not publicly available. Additionally, due to their reliance on manual annotation, all the aforementioned dictionaries have an exceedingly small number of lexicons (less than 800). This raises questions on their comprehensiveness and consequently their effectiveness for capturing textual patterns. To address these limitations, our constructed knowledge base introduces a structured lexicon of ESG concepts—one that includes multiword expressions that more effectively capture the semantics of the sustainability field, and is more extensive, comprising 23k concepts.

Additionally, existing publicly available ESG lexicons—which, to the best of our knowledge, predominantly include Baier~\citep{baier2020dict} and Naiara~\citep{Naiara2024news}—represent structured attempts to derive ESG knowledge, but differ significantly from our neuro-symbolic framework in both methodology and scalability. Baier’s dictionary is constructed through manual annotation and term frequency heuristics applied to sustainability documents, limiting its scalability and adaptability. Naiara's dictionary, by contrast, derives ESG-related terms via cosine similarity to a small set of predefined anchors (“environmental,” “social,” and “governance”), which restricts the diversity of retrieved concepts and biases term selection toward those semantically proximal to the anchors. In contrast, our approach is fully automated and designed to support both scalability and conceptual diversity. It combines GPT-4o-based annotation with label propagation, uses automated concept parsing to extract candidate ESG terms, and applies data clustering and graph-based analysis to select seed terms from semantically diverse regions of the data space. Compared to existing lexicons, our framework yields a relatively more diverse and extensive repository of ESG lexicons (or concepts), organised within a structured topical hierarchy. As highlighted by our experimental results in the later section (Section~\ref{sec:results}), this provides a greater number of unique ESG terms extracted in ESG topic analysis.

\subsection{ESG Knowledge Graphs}
While prior works have applied knowledge graphs to the analysis of sustainability data~\citep{gupta2024knowledgegraphesg, angioni2024exploringkg}, these efforts are primarily constructed to support downstream tasks such as relation extraction or question answering over ESG-related news or sustainability reports. In these cases, the knowledge graphs are designed to represent document-specific entities and relations, but are not formalized as reusable or publicly accessible repositories of ESG concepts. Moreover, these approaches are not oriented toward ESG topic analysis—they do not aim to identify, structure, or surface ESG topics and topical terms systematically across corpora. By contrast, ESGSenticNet is explicitly designed as a structured and reusable knowledge base of ESG concepts: it organizes ESG terms and their topical relations into a coherent taxonomy, and is openly available for external use. While ESGSenticNet can also support downstream applications such as relation extraction, its foundational role is to serve as a general-purpose, publicly accessible repository for ESG topic analysis. Furthermore, because ESGSenticNet can be deployed through a structured lexical method, it requires no training or technical expertise—making it accessible to stakeholders across diverse backgrounds.

\subsection{Semantic Knowledge Bases}
% what are knowledge bases and how they have been used?
In the literature, semantic knowledge bases are often represented as knowledge graphs consisting of triples in the form of (\textit{subject}, predicate, \textit{object}), or (\textit{head}, relation, \textit{tail}), which encode structural relations between entities~\citep{ji2021KGsurvey}. By structuring knowledge in a way that mimics human cognition, knowledge bases have been explored in numerous NLP applications, including commonsense question answering~\citep{dai2024commonsenseqa}, sarcasm detection~\citep{yue2023knowlenetsarcasm}, and sentiment analysis~\citep{zhong2023knowledgeabsa}, among others, as part of broader research efforts toward more context-aware and human-like language understanding. However, the construction of knowledge bases within the ESG domain remains largely unexplored. To address this gap, this paper presents the first attempt at developing a knowledge base that organises ESG knowledge into structured relationships.

% While this paper addresses this gap by developing a sustainability-focused knowledge base and leveraging it as a lexical method for topic analysis,  the structured representation of ESG concepts and their thematic relevance may also support knowledge-aware approaches in sustainability analysis.

% Semi-automated labelling methods via label propagation
% Knowledge construction can be broken down into the different sub-tasks of relation extraction, entity acquisition, triple classification, relation classification. 
Methods for constructing knowledge bases can vary according to the domain and application~\citep{ji2021KGsurvey}. Notable knowledge bases, such as ConceptNet~\citep{liu2004conceptnet} involved concept parsing and human annotation, SenticNet~\citep{senticnet} leveraged deep learning and lexical substitution, other knowledge bases involve question answering via advanced graphical neural networks (GNNs)~\citep{ye2022comprehensive}. Newer avenues for knowledge construction involve generative-LLMs, which have shown strong performance in related tasks such as reasoning, construction, and generalisation~\citep{zhu2024llmsforKB}. Unlike all the previous methods which heavily rely on semantic similarities for labelling, generative LLMs possess powerful emergent capabilities that extend beyond textual similarities~\citep{wei2022emergent}. These include reasoning, knowledge relationship construction, and generalisation to different domains~\citep{zhu2024llmsforKB}. Moreover, unlike GNNs that rely on heavily structured text data~\citep{ye2022comprehensive}, generative-LLMs can flexibly adapt to unstructured text~\cite{bernsohn2024legallensllm}. Yet, leveraging generative LLMs for knowledge construction is relatively new, with limited exploration of its effectiveness. This study contributes to this area by constructing a ESG-focused knowledge base using generative LLMs, and evaluating the resulting knowledge base's accuracy in categorising ESG concepts under the correct topics.

\subsection{Neurosymbolic Methodologies}
Neurosymbolic approaches integrate symbolic AI (rule-based systems) with sub-symbolic AI (neural networks) to enhance generalisation, interpretability, and reasoning in AI~\citep{hamilton2024neurosurvey}. These methods have been successfully applied to various NLP tasks, including word sense disambiguation for sentiment analysis~\citep{zhang2023neuro}, recommender systems~\citep{spillo2022neurorecosys}, and question-answering~\citep{dai2024commonsenseqa}. However, their potential for large-scale knowledge base construction remains largely unexplored. This study extends neurosymbolic methodologies by developing a neurosymbolic framework for knowledge construction. Specifically, sustainability linguistic patterns are integrated into rule-based concept extraction while GPT-4o is used for labelling concepts' relations with ESG topics. In doing so, this study explores the potential of a neurosymbolic approach to enhance the relevance and accuracy of knowledge construction.

\section{Conclusion}
In this study, we described the challenges of ESG topic analysis, outlining the limitations of existing NLP approaches—i.e., irrelevance, immateriality, limited organization. To specifically address these challenges, we developed a knowledge base, ESGSenticNet, from a novel neurosymbolic framework, before evaluating its effectiveness against existing NLP benchmarks. The results indicate that ESGSenticNet captures more relevant, material, and comprehensive insights for ESG topic analysis, significantly outperforming existing sustainability dictionaries and state-of-the-art NLP methods. Additionally, the topic terms ESGSenticNet provides are correctly associated with their intended topics to a reliable degree, highlighting its robustness for topic analysis. All in all, ESGSenticNet's deployment as a structured lexicon method does not require computational training or technical expertise, meaning it can be readily used by stakeholders regardless of their technical background. Ultimately, we hope that ESGSenticNet democratises access to crucial insights for ESG assessments.

% Ultimately, we hope our work democratises access to vital sustainability information, and encourages NLP development in this field. 

\section{Limitations}
Although ESGSenticNet is designed to be an accessible and inclusive tool for ESG topic analysis, it has several limitations. First, due to the evolving nature of ESG frameworks~\citep{ong2024xnlpsusanalysis}, ESGSenticNet may require periodic updates to incorporate newly emerging ESG topics. Second, the resource is currently limited to concepts expressed in English and is not yet applicable to non-English sustainability reports. Future work will therefore focus on extending ESGSenticNet to multiple languages, potentially through the use of multilingual large language models~\citep{ng2025sealion}. Third, while ESGSenticNet covers a broad range of ESG topics and is built on a large corpus of sustainability reports, it does not explicitly assess the generalisability of its concepts across reports from different industries (e.g., finance vs. energy). In other words, our study does not investigate whether ESGSenticNet provides equally effective topic analysis for sustainability reports from companies in different sectors. Addressing this may involve developing cross-industry generalisation tasks and exploring learning algorithms (i.e. contrastive learning) to improve robustness~\citep{ong2025towardsrobustesganalysis}

Additionally, while our neurosymbolic approach mitigates overreliance on LLMs (specifically GPT-4o) by incorporating sustainability-specific linguistic patterns through symbolic concept parsing, we acknowledge that LLMs may still exhibit category-specific biases due to their pretraining~\citep{dai2024bias}. For instance, they may be more accurate in classifying relations for certain sustainability categories over others~\citep{ong2025towardsrobustesganalysis}. To address this, we decompose the LLM inference task into smaller, focused sub-tasks (Section 4.3.3), improving accuracy even with less familiar categories~\citep{chang2024llmperformancesurvey}. We further validate the reliability of LLM classifications through a human evaluation study detailed in Section 5.4.

\section*{Declarations}
The authors declare that they have no competing interests as defined by Nature Portfolio, or other interests that might be perceived to influence the results and/or discussion reported in this paper. Author Contributions: K. O. : Writing – review \& editing, Writing – original draft,
Methodology, Investigation, Formal analysis, Conceptualisation. R,
M.: Writing – review \& editing, Methodology, Investigation, Concep-
tualisation. D.V.: Conceptuslisation, F.X: Conceptualisation. R. S.: Resources. J. S.: Writing – review \& editing,
Methodology, Investigation. E. C. Writing – review \& edit-
ing, Conceptualisation, Supervision. G. M.: Writing – review
\& editing, Supervision, Methodology, Conceptualisation.

\section*{Acknowledgements \& Funding Statement}
This research/project is supported by the NUS Sustainable and Green Finance Institute (SGFIN), NUS Asian Institute of Digital Finance (AIDF), Ministry of Education, Singapore under its MOE Academic
Research Fund Tier 2 (STEM RIE2025 Award MOE-T2EP20123-0005), MOE Tier 2 Award (MOE-T2EP50221-0006: ‘‘Prediction-to-Mitigation with Digital Twins of the Earth’s Weather’’), MOE Tier 1 Award (MOE-T2EP50221-0028: ‘‘Discipline-Informed Neural Networks for In-
terpretable Time-Series Discovery’’), and by the RIE2025 Industry Alignment Fund – Industry Collaboration Projects (IAF-ICP) (Award I2301E0026), administered by A*STAR, as well as supported by Alibaba Group and NTU Singapore.

\clearpage
\appendix

\section{ESGSenticNet Taxonomy}\label{appendix:taxdescr}
Table~\ref{tab:taxonomy} contains the definitions of the pillars (Environmental, Social, Governance) and topics used in ESGSenticNet. Along with the pillars, the broad topics, and cross-broad topics are bolded, with their constituent sub-topics highlighted in bullet points. Beside the broad and cross-broad topic names, a letter denotes which pillar the topic falls under -- (E) Environmental, (S) Social, (G) Governance.

\begin{table}[H]
\centering
\small
\begin{tabularx}{\textwidth}{|X|X|}
\hline
\textbf{Topic} & \textbf{Description} \\
\hline
\textbf{Environmental} & Focuses on impact on the environment. This includes, but is not limited to, green innovation, green infrastructure, resource stewardship, energy consumption, waste management, pollution reduction, natural resource conservation, and animal welfare.\\
\hline
\textbf{Social} & Focuses on managing relationships with its employees, suppliers, customers, and the communities. This includes, but is not limited to, employee diversity, equity, working conditions, customer relations, partnerships and social engagement.\\
\hline
\textbf{Governance} & Focuses on the system of rules, practices, and processes by which a company is directed and controlled. This includes, but is not limited to, corporate oversight, corporate transparency, executive compensation, ethical behavior, and shareholder rights. \\
\hline
\textbf{Development (E)} & Drives innovation and growth in sustainability by creating new technologies or infrastructure for sustainability, or supporting these efforts through green financing. \\
\hline
\quad \scalebox{0.7}{$\bullet$} Green Technologies & Technological systems, products, and processes that assist in reducing environmental impact. \\
\hline
\quad \scalebox{0.7}{$\bullet$} Green Infrastructure & Environmentally sustainable buildings, infrastructure, energy solutions and urban systems. \\
\hline
\quad \scalebox{0.7}{$\bullet$} Green Financing & Financial services that support efforts aimed at environmental sustainability. \\
\hline
\textbf{Ecological Conservation (E)} & Manages the company’s interaction with natural ecosystems, focusing on minimizing harmful impacts and promoting biodiversity and habitat conservation. \\
\hline
\end{tabularx}
\caption{ESGSenticNet Sustainability Taxonomy - continued in the next page}
\end{table} 

\begin{table}[H]
\ContinuedFloat
\small
\centering
\begin{tabularx}{\textwidth}{|X|X|}
\hline
\textbf{Emissions Control (E)} & Targets the reduction of various emissions from company operations, with an emphasis on mitigating climate change impacts or protecting air quality. \\
\hline
\quad \scalebox{0.7}{$\bullet$} Climate Emissions & Manages greenhouse gas emissions that contribute to global warming, such as CO2 and methane. The phrase must explicitly relate to greenhouse gases or related terms such as 'carbon', 'methane'. \\
\hline
\quad \scalebox{0.7}{$\bullet$} Air Quality Emissions & Manages emissions of non-greenhouse gases, like toxic gas, particulates and sulfur dioxide, which are critical for local air quality and public health. \\
\hline
\textbf{Operations (E)} & Focuses on optimizing existing business activities for the sole purpose of minimizing environmental impact. This includes minimising environmental impact by enhancing current practices in corporate operations, supply chains, production processes, or land management, distinct from the development of innovative solutions and technologies. \\
\hline
\quad \scalebox{0.7}{$\bullet$} Sustainable Corporate Operations & Environmentally responsible practices within corporate offices. Examples include but are not limited to promoting sustainability amongst employees, eco-friendly office lighting.   \\
\hline
\quad \scalebox{0.7}{$\bullet$} Sustainable Supply Chain & Logistics and distribution methods or practices that are environmentally responsible, ensuring sustainability in the movement and handling of goods. \\
\hline
\quad \scalebox{0.7}{$\bullet$} Sustainable Production Processes & Manufacturing processes or practices that are environmentally responsible. \\
\hline
\quad \scalebox{0.7}{$\bullet$} Sustainable Land Management & Using and managing land in a way that is environmentally responsible, conserves natural resources and protects biodiversity.  \\
\hline
\textbf{Waste Management (E)} & Concentrates on managing waste after it has been produced, including the reduction, handling, disposal, and treatment of solid waste and wastewater. It involves but is not limited to recycling, reusing materials, and employing wastewater treatment processes to mitigate environmental degradation and facilitate resource recovery. \\
\hline
\quad \scalebox{0.7}{$\bullet$} Wastewater Management & Focuses on the treatment, reuse and recycling of wastewater or liquid waste to prevent pollution or conserve water resources. \\
\hline
\end{tabularx}
\caption{ESGSenticNet Sustainability Taxonomy - continued in the next page}
\end{table}

\begin{table}[H]
\ContinuedFloat
\small
\centering  % Centers the content (the table) within the minipage  % Adjusts the vertical space at the top of the page
\begin{tabularx}{\textwidth}{|X|X|}
\hline
\quad \scalebox{0.7}{$\bullet$} Solid Waste Management & Addresses the management of solid wastes, promoting reduction, recycling treatment, and responsible disposal to minimize environmental impact and encourage circular economy principles.\\
\hline
\textbf{Resource Optimisation (E)} & Prioritizes the proactive, efficient, and sustainable selection and use of energy, water, and materials. This approach focuses on minimizing environmental impact through the strategic choice and utilization of resources. Distinct from waste management as it does not primarily deal with waste but rather prevents waste generation by optimizing resource use from the start. \\
\hline
\quad \scalebox{0.7}{$\bullet$} Renewable and Efficient Energy & Focuses on using energy efficiently, reducing energy consumption and increasing the use of renewable energy sources such as solar, wind, and hydroelectric power to minimize environmental impact. \\
\hline
\quad \scalebox{0.7}{$\bullet$} Material Sustainability & Prioritizes the use of sustainable and renewable materials, focusing on maximizing their efficiency from the outset. Distinct from recycling and reuse, which are part of waste management, this approach aims to reduce environmental impact before materials enter the waste stream. \\
\hline
\quad \scalebox{0.7}{$\bullet$} Water Conservation & Focuses on using water efficienctly, reducing water usage and enhancing sustainable water management practices to minimize initial consumption. Distinct from water recycling and reusing, which address management of wastewater. \\
\hline
\textbf{Workplace (S)} & Focuses on the social aspects of sustainability within the company to promote equity, inclusivity, or quality of life, emphasizing employee well-being, inclusivity, or professional growth. \\
\hline
\quad \scalebox{0.7}{$\bullet$} Workplace Wellness & Efforts that focus on the satisfaction, emotional and physical well-being of employees. \\
\hline
\quad \scalebox{0.7}{$\bullet$} Workplace Diversity & Efforts that emphasise gender and racial diversity and fairness in the workplace, such as anti-discrimination measures, fair hiring practices etc.  \\
\hline
\quad \scalebox{0.7}{$\bullet$} Workplace Development & Efforts that focus on the skill training and professional development of staff. \\
\hline
\end{tabularx}
\caption{ESGSenticNet Sustainability Taxonomy - continued in the next page}
\end{table}

\begin{table}[H]
\ContinuedFloat
\small
\centering  % Centers the content (the table) within the minipage  % Adjusts the vertical space at the top of the page
\begin{tabularx}{\textwidth}{|X|X|}
\hline
\textbf{Communications (E/S/G)} & Focuses on how a company transparently and honestly communicates and discloses its activities. \\
\hline
\textbf{Outreach (S)} & Enhances social sustainability by contributing to community well-being, fostering customer satisfaction, building customer relationships, or collaborating with strategic partners to promote equity, inclusivity, and quality of life. \\
\hline
\quad \scalebox{0.7}{$\bullet$} Community Empowerment & Contributions to the welfare of local communities through various forms of support and engagement.  \\
\hline
\quad \scalebox{0.7}{$\bullet$} Strategic Partnerships & Collaborations with stakeholders like NGOs, businesses, and charities to achieve shared sustainability goals.\\
\hline
\quad \scalebox{0.7}{$\bullet$} Customer Engagement & Fostering customer satisfaction, strong relationships and reputation with customers.\\
\hline
\textbf{Management (G)} & Ensures strategic, ethical, or effective governance through rigorous oversight, diverse board composition, active stakeholder engagement, or high ethical standards within company leadership. \\
\hline
\quad \scalebox{0.7}{$\bullet$} Management Composition & Structures the board or management with diverse expertise and backgrounds to strengthen decision-making and governance effectiveness. \\
\hline
\quad \scalebox{0.7}{$\bullet$} Management Ethics & Enforces strict ethical standards within company leadership, ensuring management actions uphold corporate integrity. \\
\hline
\quad \scalebox{0.7}{$\bullet$} Management Compensation & Aligns management, executive and board remuneration with company performance, sustainability or ethical objectives. \\
\hline
\quad \scalebox{0.7}{$\bullet$} Stakeholder Engagement & Engages key stakeholders to align board or management decisions with broader interests and feedback. \\
\hline
\quad \scalebox{0.7}{$\bullet$} Oversight & Involves the systematic review and evaluation of management actions to ensure they align with the company's performance goals, sustainability goals and ethical standards. \\
\hline
\textbf{Compliance (G)} & Ensures adherence to laws, regulations, or standards across environmental protection, labor rights, consumer safety, ethical conduct, financial practices, or data protection, distinguishing itself from voluntary sustainability initiatives by focusing on mandatory requirements. \\
\hline
\end{tabularx}
\caption{ESGSenticNet Sustainability Taxonomy - continued in the next page}
\end{table}

\begin{table}[H]
\ContinuedFloat
\small
\centering  % Centers the content (the table) within the minipage  % Adjusts the vertical space at the top of the page
\begin{tabularx}{\textwidth}{|X|X|}
\hline
\quad \scalebox{0.7}{$\bullet$} Environmental Compliance & Adherence to environmental laws such as for emissions, water pollution, protection of biodiversity etc. \\
\hline
\quad \scalebox{0.7}{$\bullet$} Worker \& Consumer Safety & Adherence to laws that guarantee the safety of products and services to consumer health, as well as fair labor practices and the protection of workers' rights. \\
\hline
\quad \scalebox{0.7}{$\bullet$} Business Compliance & Adherence to anti-corruption, anti-competitive practices, financial regulations, or financial reporting. \\
\hline
\quad \scalebox{0.7}{$\bullet$} Data Privacy \& Cybersecurity Protection & Protection and confidentiality of data, which includes measures against cyber threats and compliance with data security regulations. \\
\hline
% Add other rows as needed
\end{tabularx}
\caption{ESGSenticNet Sustainability Taxonomy}
\label{tab:taxonomy}
\end{table} 

\section{Additional details on Human Validation}\label{sec:humanannotate}
For evaluating ESGSenticNet accuracy (Section~\ref{sec:evaluation}), ESGSenticNet's flexible matching method (Section~\ref{sec:flexible}), GPT-4 human agreement (Section~\ref{sec:agreement}), there are a total of 9 human annotators. These annotators are based in Singapore, comprising researchers from the Asian Institute of Digital Finance and the Sustainable and Green Finance Institute, who are pursuing doctoral-level and post-doctoral research within the corporate sustainability field. All human annotators contribute to this study as part of their formal academic and research responsibilities. All human validation studies follow the same annotation scheme. When there are disagreements, discussions are held to achieve a resolution. Where there is no clear resolution, majority voting is taken to determine the ground truth label. For the derivation of the taxonomy for ESGSenticNet, we involved two tenured researchers from the Singapore Sustainable and Green Finance Institute (SGFIN), with expertise within the sustainable finance field.

\section{Full Statistical Breakdown of ESGSenticNet}\label{appendix:c}

\begin{singlespace}
\begin{table}[H]
\centering
\resizebox{\textwidth}{!}{%
\begin{tabular}{|p{6.8cm}|p{1.6cm}|p{2cm}|p{2cm}|p{2cm}|}
\hline
\textbf{Topics}  &\textbf{Total triples} & \multicolumn{3}{c|}{\textbf{Number of triples with the relations}}\\ 
\cline{3-5}
& & \textbf{supports} & 
\textbf{undermines} & \textbf{aligns with} \\
\hline
% \textbf{Aligns with} & 23245 & - & - & - \\
% \hline
% \textbf{Supports} & 19414 & - & - & - \\
% \hline
% \textbf{Undermines} & 1573 & - & - & - \\
% \hline
\textbf{Environmental} & 5904 & - & - & 5904\\
\hline
\textbf{Social} & 6897 & - & - & 6897 \\
\hline
\textbf{Governance} & 10444 & - & - & 10444 \\
\hline
\textbf{Development (E)} & 314 & 298 & 16 & \\
\hline
\quad \scalebox{0.7}{$\bullet$} Green Technologies & 58 & 49 & 9 & - \\
\hline
\quad \scalebox{0.7}{$\bullet$} Green Infrastructure & 105 & 104 & 1 & - \\
\hline
\quad \scalebox{0.7}{$\bullet$} Green Financing & 16 & 16 & - & - \\
\hline
\textbf{Operations (E)} & 1470 & 1209 & 261 & - \\
\hline
\quad \scalebox{0.7}{$\bullet$} Sustainable Corporate Operations & 215 & 186 & 29 & - \\
\hline
\quad \scalebox{0.7}{$\bullet$} Sustainable Supply Chain & 48 & 46 & 2 & - \\
\hline
\quad \scalebox{0.7}{$\bullet$} Sustainable Production Processes & 265 & 148 & 117 & - \\
\hline
\quad \scalebox{0.7}{$\bullet$} Sustainable Land Management & 174 & 141 & 33 & - \\
\hline
\textbf{Resource Optimisation (E)} & 935 & 855 & 80 & - \\
\hline
\quad \scalebox{0.7}{$\bullet$} Renewable and Efficient Energy & 348 & 321 & 27 & - \\
\hline
\quad \scalebox{0.7}{$\bullet$} Material Sustainability & 58 & 50 & 8 & - \\
\hline
\quad \scalebox{0.7}{$\bullet$} Water Conservation & 107 & 100 & 7 & - \\
\hline
\textbf{Waste Management (E)} & 578 & 493 & 83 & - \\
\hline
\quad \scalebox{0.7}{$\bullet$} Wastewater Management & 82 & 67 & 15 & - \\
\hline
\quad \scalebox{0.7}{$\bullet$} Solid Waste Management & 166 & 156 & 10 & - \\
\hline
\textbf{Emissions Control (E)} & 533 & 407 & 146 & - \\
\hline
\quad \scalebox{0.7}{$\bullet$} Climate Emissions & 173 & 152 & 21 & - \\
\hline
\quad \scalebox{0.7}{$\bullet$} Air Quality Emissions & 36 & 19 & 17 & \\
\hline
\textbf{Ecological Conservation (E)} & 238 & 205 & 33 & - \\
\hline
\textbf{Workplace (S)} & 2077 & 2006 & 71 & - \\
\hline
\quad \scalebox{0.7}{$\bullet$} Workplace Wellness & 463 & 432 & 31 & - \\
\hline
\quad \scalebox{0.7}{$\bullet$} Workplace Diversity & 237 & 222 & 15 & - \\
\hline
\quad \scalebox{0.7}{$\bullet$} Workplace Development & 705 & 704 & 1 & - \\
\hline
\textbf{Outreach (S)} & 1212 & 1211 & 1 & - \\
\hline
\quad \scalebox{0.7}{$\bullet$} Community Empowerment & 498 & 498 & 0 & - \\
\hline
\quad \scalebox{0.7}{$\bullet$} Strategic Partnerships & 137 & 137 & 0 & - \\
\hline
\quad \scalebox{0.7}{$\bullet$} Customer Engagement & 77 & 77 & 0 & - \\
\hline
\textbf{Management (G)} & 2332 & 2305 & 27 & - \\
\hline
\quad \scalebox{0.7}{$\bullet$} Management Composition & 30 & 30 & - & - \\
\hline
\quad \scalebox{0.7}{$\bullet$} Management Ethics & 124 & 118 & 6 & - \\
\hline
\quad \scalebox{0.7}{$\bullet$} Management Compensation & 19 & 17 & 2 & - \\
\hline
\quad \scalebox{0.7}{$\bullet$} Stakeholder Engagement & 105 & 105 & - & - \\
\hline
\quad \scalebox{0.7}{$\bullet$} Oversight & 532 & 532 & - & - \\
\hline
\textbf{Communications (E/S/G)} & 1196 & 1192 & 4 & - \\
\hline
\textbf{Compliance (G)} & 3405 & 3093 & 312 & - \\
\hline
\quad \scalebox{0.7}{$\bullet$} Environmental Compliance & 436 & 403 & 33 & - \\
\hline
\quad \scalebox{0.7}{$\bullet$} Worker \& Consumer Safety & 1154 & 1040 & 114 & - \\
\hline
\quad \scalebox{0.7}{$\bullet$} Business Compliance & 208 & 170 & 38 & - \\
\hline
\quad \scalebox{0.7}{$\bullet$} Data Privacy \& Cybersecurity Protection & 103 & 100 & 3 & - \\
\hline
\end{tabular}%
}
\caption{ESGSenticNet Statistics}
\label{tab:stats}
\end{table}
\end{singlespace}

% \end{lineo}
% \end{linenumbers}

\clearpage

\bibliographystyle{elsarticle-harv}

\bibliography{main}

\begin{thebibliography}{83}
\expandafter\ifx\csname natexlab\endcsname\relax\def\natexlab#1{#1}\fi
\providecommand{\url}[1]{\texttt{#1}}
\providecommand{\href}[2]{#2}
\providecommand{\path}[1]{#1}
\providecommand{\DOIprefix}{doi:}
\providecommand{\ArXivprefix}{arXiv:}
\providecommand{\URLprefix}{URL: }
\providecommand{\Pubmedprefix}{pmid:}
\providecommand{\doi}[1]{\href{http://dx.doi.org/#1}{\path{#1}}}
\providecommand{\Pubmed}[1]{\href{pmid:#1}{\path{#1}}}
\providecommand{\bibinfo}[2]{#2}
\ifx\xfnm\relax \def\xfnm[#1]{\unskip,\space#1}\fi
%Type = Article
\bibitem[{Abdelrazek et~al.(2023)Abdelrazek, Eid, Gawish, Medhat and Hassan}]{abdelrazek2023topicsurvey2}
\bibinfo{author}{Abdelrazek, A.}, \bibinfo{author}{Eid, Y.}, \bibinfo{author}{Gawish, E.}, \bibinfo{author}{Medhat, W.}, \bibinfo{author}{Hassan, A.}, \bibinfo{year}{2023}.
\newblock \bibinfo{title}{Topic modeling algorithms and applications: A survey}.
\newblock \bibinfo{journal}{Information Systems} \bibinfo{volume}{112}, \bibinfo{pages}{102131}.
%Type = Article
\bibitem[{Abram et~al.(2020)Abram, Mancini and Parker}]{abram2020methods}
\bibinfo{author}{Abram, M.D.}, \bibinfo{author}{Mancini, K.T.}, \bibinfo{author}{Parker, R.D.}, \bibinfo{year}{2020}.
\newblock \bibinfo{title}{Methods to integrate natural language processing into qualitative research}.
\newblock \bibinfo{journal}{International Journal of Qualitative Methods} \bibinfo{volume}{19}, \bibinfo{pages}{1609406920984608}.
%Type = Inproceedings
\bibitem[{Aggarwal et~al.(2001)Aggarwal, Hinneburg and Keim}]{aggarwal2001surprisingdistance}
\bibinfo{author}{Aggarwal, C.C.}, \bibinfo{author}{Hinneburg, A.}, \bibinfo{author}{Keim, D.A.}, \bibinfo{year}{2001}.
\newblock \bibinfo{title}{On the surprising behavior of distance metrics in high dimensional space}, in: \bibinfo{booktitle}{Database theory---ICDT 2001: 8th international conference London, UK, January 4--6, 2001 proceedings 8}, \bibinfo{organization}{Springer}. pp. \bibinfo{pages}{420--434}.
%Type = Article
\bibitem[{Angioni et~al.(2024)Angioni, Consoli, Dess{\'\i}, Osborne, Recupero and Salatino}]{angioni2024exploringkg}
\bibinfo{author}{Angioni, S.}, \bibinfo{author}{Consoli, S.}, \bibinfo{author}{Dess{\'\i}, D.}, \bibinfo{author}{Osborne, F.}, \bibinfo{author}{Recupero, D.R.}, \bibinfo{author}{Salatino, A.}, \bibinfo{year}{2024}.
\newblock \bibinfo{title}{Exploring environmental, social, and governance (esg) discourse in news: An ai-powered investigation through knowledge graph analysis}.
\newblock \bibinfo{journal}{IEEE Access} .
%Type = Article
\bibitem[{Baier et~al.(2020)Baier, Berninger and Kiesel}]{baier2020dict}
\bibinfo{author}{Baier, P.}, \bibinfo{author}{Berninger, M.}, \bibinfo{author}{Kiesel, F.}, \bibinfo{year}{2020}.
\newblock \bibinfo{title}{Environmental, social and governance reporting in annual reports: A textual analysis}.
\newblock \bibinfo{journal}{Financial Markets, Institutions \& Instruments} \bibinfo{volume}{29}, \bibinfo{pages}{93--118}.
%Type = Article
\bibitem[{Berg et~al.(2022)Berg, Koelbel and Rigobon}]{berg2022aggregate}
\bibinfo{author}{Berg, F.}, \bibinfo{author}{Koelbel, J.F.}, \bibinfo{author}{Rigobon, R.}, \bibinfo{year}{2022}.
\newblock \bibinfo{title}{Aggregate confusion: The divergence of esg ratings}.
\newblock \bibinfo{journal}{Review of Finance} \bibinfo{volume}{26}, \bibinfo{pages}{1315--1344}.
%Type = Inproceedings
\bibitem[{Bernsohn et~al.(2024)Bernsohn, Semo, Vazana, Hayat, Hagag, Niklaus, Saha and Truskovskyi}]{bernsohn2024legallensllm}
\bibinfo{author}{Bernsohn, D.}, \bibinfo{author}{Semo, G.}, \bibinfo{author}{Vazana, Y.}, \bibinfo{author}{Hayat, G.}, \bibinfo{author}{Hagag, B.}, \bibinfo{author}{Niklaus, J.}, \bibinfo{author}{Saha, R.}, \bibinfo{author}{Truskovskyi, K.}, \bibinfo{year}{2024}.
\newblock \bibinfo{title}{{L}egal{L}ens: Leveraging {LLM}s for legal violation identification in unstructured text}, in: \bibinfo{booktitle}{Proceedings of the 18th Conference of the European Chapter of the Association for Computational Linguistics (Volume 1: Long Papers)}, \bibinfo{publisher}{Association for Computational Linguistics}. pp. \bibinfo{pages}{2129--2145}.
%Type = Article
\bibitem[{Beske et~al.(2020)Beske, Haustein and Lorson}]{beske2020sustainabilityactionsmaterial}
\bibinfo{author}{Beske, F.}, \bibinfo{author}{Haustein, E.}, \bibinfo{author}{Lorson, P.C.}, \bibinfo{year}{2020}.
\newblock \bibinfo{title}{Materiality analysis in sustainability and integrated reports}.
\newblock \bibinfo{journal}{Sustainability Accounting, Management and Policy Journal} \bibinfo{volume}{11}, \bibinfo{pages}{162--186}.
%Type = Article
\bibitem[{Bi et~al.(2023)Bi, Guo, Zhao, Sun and Wang}]{bi2023lexical}
\bibinfo{author}{Bi, D.}, \bibinfo{author}{Guo, J.e.}, \bibinfo{author}{Zhao, E.}, \bibinfo{author}{Sun, S.}, \bibinfo{author}{Wang, S.}, \bibinfo{year}{2023}.
\newblock \bibinfo{title}{Using word embedding for environmental violation analysis: Evidence from pennsylvania unconventional oil and gas compliance reports}.
\newblock \bibinfo{journal}{Environmental Development} \bibinfo{volume}{47}, \bibinfo{pages}{100905}.
%Type = Article
\bibitem[{Blei et~al.(2003)Blei, Ng and Jordan}]{blei2003lda}
\bibinfo{author}{Blei, D.M.}, \bibinfo{author}{Ng, A.Y.}, \bibinfo{author}{Jordan, M.I.}, \bibinfo{year}{2003}.
\newblock \bibinfo{title}{Latent dirichlet allocation}.
\newblock \bibinfo{journal}{Journal of machine Learning research} \bibinfo{volume}{3}, \bibinfo{pages}{993--1022}.
%Type = Article
\bibitem[{Bochkay et~al.(2023)Bochkay, Brown, Leone and Tucker}]{bochkay2023textualaccounting}
\bibinfo{author}{Bochkay, K.}, \bibinfo{author}{Brown, S.V.}, \bibinfo{author}{Leone, A.J.}, \bibinfo{author}{Tucker, J.W.}, \bibinfo{year}{2023}.
\newblock \bibinfo{title}{Textual analysis in accounting: What's next?}
\newblock \bibinfo{journal}{Contemporary accounting research} \bibinfo{volume}{40}, \bibinfo{pages}{765--805}.
%Type = Article
\bibitem[{Camana et~al.(2021)Camana, Manzardo, Toniolo, Gallo and Scipioni}]{camana2021envpolicyanalysis}
\bibinfo{author}{Camana, D.}, \bibinfo{author}{Manzardo, A.}, \bibinfo{author}{Toniolo, S.}, \bibinfo{author}{Gallo, F.}, \bibinfo{author}{Scipioni, A.}, \bibinfo{year}{2021}.
\newblock \bibinfo{title}{Assessing environmental sustainability of local waste management policies in italy from a circular economy perspective. an overview of existing tools}.
\newblock \bibinfo{journal}{Sustainable Production and Consumption} \bibinfo{volume}{27}, \bibinfo{pages}{613--629}.
%Type = Inproceedings
\bibitem[{Cambria et~al.(2022)Cambria, Mao, Han and Liu}]{cambria2022sentic}
\bibinfo{author}{Cambria, E.}, \bibinfo{author}{Mao, R.}, \bibinfo{author}{Han, S.}, \bibinfo{author}{Liu, Q.}, \bibinfo{year}{2022}.
\newblock \bibinfo{title}{Sentic parser: A graph-based approach to concept extraction for sentiment analysis}, in: \bibinfo{booktitle}{IEEE International Conference on Data Mining (ICDM)}, \bibinfo{organization}{IEEE}. pp. \bibinfo{pages}{413--420}.
%Type = Inproceedings
\bibitem[{Cambria et~al.(2024)Cambria, Zhang, Mao, Chen and Kwok}]{senticnet}
\bibinfo{author}{Cambria, E.}, \bibinfo{author}{Zhang, X.}, \bibinfo{author}{Mao, R.}, \bibinfo{author}{Chen, M.}, \bibinfo{author}{Kwok, K.}, \bibinfo{year}{2024}.
\newblock \bibinfo{title}{{SenticNet} 8: Fusing emotion ai and commonsense ai for interpretable, trustworthy, and explainable affective computing}, in: \bibinfo{booktitle}{International Conference on Human-Computer Interaction (HCII)}, \bibinfo{address}{Washington DC, USA}. pp. \bibinfo{pages}{197--216}.
%Type = Inproceedings
\bibitem[{Campello et~al.(2013)Campello, Moulavi and Sander}]{campello2013hdbscan}
\bibinfo{author}{Campello, R.J.}, \bibinfo{author}{Moulavi, D.}, \bibinfo{author}{Sander, J.}, \bibinfo{year}{2013}.
\newblock \bibinfo{title}{Density-based clustering based on hierarchical density estimates}, in: \bibinfo{booktitle}{Pacific-Asia conference on knowledge discovery and data mining}, \bibinfo{organization}{Springer}. pp. \bibinfo{pages}{160--172}.
%Type = Article
\bibitem[{Cenci et~al.(2023)Cenci, Burato, Rei and Zollo}]{cenci2023corporatesusimpt}
\bibinfo{author}{Cenci, S.}, \bibinfo{author}{Burato, M.}, \bibinfo{author}{Rei, M.}, \bibinfo{author}{Zollo, M.}, \bibinfo{year}{2023}.
\newblock \bibinfo{title}{The alignment of companies' sustainability behavior and emissions with global climate targets}.
\newblock \bibinfo{journal}{Nature Communications} \bibinfo{volume}{14}, \bibinfo{pages}{7831}.
%Type = Article
\bibitem[{Chang et~al.(2024)Chang, Wang, Wang, Wu, Yang, Zhu, Chen, Yi, Wang, Wang et~al.}]{chang2024llmperformancesurvey}
\bibinfo{author}{Chang, Y.}, \bibinfo{author}{Wang, X.}, \bibinfo{author}{Wang, J.}, \bibinfo{author}{Wu, Y.}, \bibinfo{author}{Yang, L.}, \bibinfo{author}{Zhu, K.}, \bibinfo{author}{Chen, H.}, \bibinfo{author}{Yi, X.}, \bibinfo{author}{Wang, C.}, \bibinfo{author}{Wang, Y.}, et~al., \bibinfo{year}{2024}.
\newblock \bibinfo{title}{A survey on evaluation of large language models}.
\newblock \bibinfo{journal}{ACM Transactions on Intelligent Systems and Technology} \bibinfo{volume}{15}, \bibinfo{pages}{1--45}.
%Type = Article
\bibitem[{Chiang and Lee(2023)}]{chiang2023llmhuman}
\bibinfo{author}{Chiang, C.H.}, \bibinfo{author}{Lee, H.y.}, \bibinfo{year}{2023}.
\newblock \bibinfo{title}{Can large language models be an alternative to human evaluations?}
\newblock \bibinfo{journal}{arXiv preprint arXiv:2305.01937} .
%Type = Article
\bibitem[{Constant et~al.(2017)Constant, Eryi{\u{g}}it, Monti, Van Der~Plas, Ramisch, Rosner and Todirascu}]{constant2017multiword}
\bibinfo{author}{Constant, M.}, \bibinfo{author}{Eryi{\u{g}}it, G.}, \bibinfo{author}{Monti, J.}, \bibinfo{author}{Van Der~Plas, L.}, \bibinfo{author}{Ramisch, C.}, \bibinfo{author}{Rosner, M.}, \bibinfo{author}{Todirascu, A.}, \bibinfo{year}{2017}.
\newblock \bibinfo{title}{Multiword expression processing: A survey}.
\newblock \bibinfo{journal}{Computational Linguistics} \bibinfo{volume}{43}, \bibinfo{pages}{837--892}.
%Type = Inproceedings
\bibitem[{Dai et~al.(2024a)Dai, Tan, Mo, Liang, Huo, Luo and Cheng}]{dai2024commonsenseqa}
\bibinfo{author}{Dai, R.}, \bibinfo{author}{Tan, Y.}, \bibinfo{author}{Mo, L.}, \bibinfo{author}{Liang, S.}, \bibinfo{author}{Huo, G.}, \bibinfo{author}{Luo, J.}, \bibinfo{author}{Cheng, Y.}, \bibinfo{year}{2024}a.
\newblock \bibinfo{title}{G-sap: Graph-based structure-aware prompt learning over heterogeneous knowledge for commonsense reasoning}, in: \bibinfo{booktitle}{Proceedings of the 2024 International Conference on Multimedia Retrieval}, pp. \bibinfo{pages}{1051--1060}.
%Type = Inproceedings
\bibitem[{Dai et~al.(2024b)Dai, Xu, Xu, Pang, Dong and Xu}]{dai2024bias}
\bibinfo{author}{Dai, S.}, \bibinfo{author}{Xu, C.}, \bibinfo{author}{Xu, S.}, \bibinfo{author}{Pang, L.}, \bibinfo{author}{Dong, Z.}, \bibinfo{author}{Xu, J.}, \bibinfo{year}{2024}b.
\newblock \bibinfo{title}{Bias and unfairness in information retrieval systems: New challenges in the llm era}, in: \bibinfo{booktitle}{Proceedings of the 30th ACM SIGKDD Conference on Knowledge Discovery and Data Mining}, pp. \bibinfo{pages}{6437--6447}.
%Type = Article
\bibitem[{Derqui(2020)}]{derqui2020towardsevolving}
\bibinfo{author}{Derqui, B.}, \bibinfo{year}{2020}.
\newblock \bibinfo{title}{Towards sustainable development: Evolution of corporate sustainability in multinational firms}.
\newblock \bibinfo{journal}{Corporate Social Responsibility and Environmental Management} \bibinfo{volume}{27}, \bibinfo{pages}{2712--2723}.
%Type = Article
\bibitem[{Du et~al.(2024)Du, Bao, Meng and Hui}]{du2024esgbigramtext}
\bibinfo{author}{Du, F.}, \bibinfo{author}{Bao, C.}, \bibinfo{author}{Meng, Q.}, \bibinfo{author}{Hui, Y.}, \bibinfo{year}{2024}.
\newblock \bibinfo{title}{Text-based measure of esg risk exposure}.
\newblock \bibinfo{journal}{Procedia Computer Science} \bibinfo{volume}{242}, \bibinfo{pages}{693--700}.
%Type = Inproceedings
\bibitem[{Du et~al.(2023)Du, Xing, Mao and Cambria}]{du2023finsenticnet}
\bibinfo{author}{Du, K.}, \bibinfo{author}{Xing, F.}, \bibinfo{author}{Mao, R.}, \bibinfo{author}{Cambria, E.}, \bibinfo{year}{2023}.
\newblock \bibinfo{title}{{FinSenticNet}: A concept-level lexicon for financial sentiment analysis}, in: \bibinfo{booktitle}{IEEE Symposium Series on Computational Intelligence (SSCI)}, \bibinfo{organization}{IEEE}. pp. \bibinfo{pages}{109--114}.
%Type = Article
\bibitem[{Eng et~al.(2022)Eng, Fikru and Vichitsarawong}]{eng2022sasb}
\bibinfo{author}{Eng, L.L.}, \bibinfo{author}{Fikru, M.}, \bibinfo{author}{Vichitsarawong, T.}, \bibinfo{year}{2022}.
\newblock \bibinfo{title}{Comparing the informativeness of sustainability disclosures versus esg disclosure ratings}.
\newblock \bibinfo{journal}{Sustainability Accounting, Management and Policy Journal} \bibinfo{volume}{13}, \bibinfo{pages}{494--518}.
%Type = Article
\bibitem[{Eslami et~al.(2021)Eslami, Lezoche, Panetto and Dassisti}]{eslami2021analysingframework2}
\bibinfo{author}{Eslami, Y.}, \bibinfo{author}{Lezoche, M.}, \bibinfo{author}{Panetto, H.}, \bibinfo{author}{Dassisti, M.}, \bibinfo{year}{2021}.
\newblock \bibinfo{title}{On analysing sustainability assessment in manufacturing organisations: a survey}.
\newblock \bibinfo{journal}{International Journal of Production Research} \bibinfo{volume}{59}, \bibinfo{pages}{4108--4139}.
%Type = Article
\bibitem[{Ferjan{\v{c}}i{\v{c}} et~al.(2024a)Ferjan{\v{c}}i{\v{c}}, Ichev, Lon{\v{c}}arski, Montariol, Pelicon, Pollak, {\v{S}}u{\v{s}}tar, Toman, Valentin{\v{c}}i{\v{c}} and {\v{Z}}nidar{\v{s}}i{\v{c}}}]{ferjanvcivc2024textualanalysis}
\bibinfo{author}{Ferjan{\v{c}}i{\v{c}}, U.}, \bibinfo{author}{Ichev, R.}, \bibinfo{author}{Lon{\v{c}}arski, I.}, \bibinfo{author}{Montariol, S.}, \bibinfo{author}{Pelicon, A.}, \bibinfo{author}{Pollak, S.}, \bibinfo{author}{{\v{S}}u{\v{s}}tar, K.S.}, \bibinfo{author}{Toman, A.}, \bibinfo{author}{Valentin{\v{c}}i{\v{c}}, A.}, \bibinfo{author}{{\v{Z}}nidar{\v{s}}i{\v{c}}, M.}, \bibinfo{year}{2024}a.
\newblock \bibinfo{title}{Textual analysis of corporate sustainability reporting and corporate esg scores}.
\newblock \bibinfo{journal}{International Review of Financial Analysis} \bibinfo{volume}{96}, \bibinfo{pages}{103669}.
%Type = Article
\bibitem[{Ferjan{\v{c}}i{\v{c}} et~al.(2024b)Ferjan{\v{c}}i{\v{c}}, Ichev, Lon{\v{c}}arski, Montariol, Pelicon, Pollak, {\v{S}}u{\v{s}}tar, Toman, Valentin{\v{c}}i{\v{c}} and {\v{Z}}nidar{\v{s}}i{\v{c}}}]{ferjanvcivc2024textualbertopicesg}
\bibinfo{author}{Ferjan{\v{c}}i{\v{c}}, U.}, \bibinfo{author}{Ichev, R.}, \bibinfo{author}{Lon{\v{c}}arski, I.}, \bibinfo{author}{Montariol, S.}, \bibinfo{author}{Pelicon, A.}, \bibinfo{author}{Pollak, S.}, \bibinfo{author}{{\v{S}}u{\v{s}}tar, K.S.}, \bibinfo{author}{Toman, A.}, \bibinfo{author}{Valentin{\v{c}}i{\v{c}}, A.}, \bibinfo{author}{{\v{Z}}nidar{\v{s}}i{\v{c}}, M.}, \bibinfo{year}{2024}b.
\newblock \bibinfo{title}{Textual analysis of corporate sustainability reporting and corporate esg scores}.
\newblock \bibinfo{journal}{International Review of Financial Analysis} \bibinfo{volume}{96}, \bibinfo{pages}{103669}.
%Type = Misc
\bibitem[{Grootendorst(2020)}]{grootendorst2020keybert}
\bibinfo{author}{Grootendorst, M.}, \bibinfo{year}{2020}.
\newblock \bibinfo{title}{{KeyBERT: Minimal keyword extraction with BERT}}.
\newblock \URLprefix \url{https://doi.org/10.5281/zenodo.4461265}.
%Type = Article
\bibitem[{Grootendorst(2022)}]{grootendorst2022bertopic}
\bibinfo{author}{Grootendorst, M.}, \bibinfo{year}{2022}.
\newblock \bibinfo{title}{Bertopic: Neural topic modeling with a class-based tf-idf procedure}.
\newblock \bibinfo{journal}{arXiv preprint arXiv:2203.05794} .
%Type = Article
\bibitem[{Gupta et~al.(2024)Gupta, Goel, Verma, Dey and Bhardwaj}]{gupta2024knowledgegraphesg}
\bibinfo{author}{Gupta, T.K.}, \bibinfo{author}{Goel, T.}, \bibinfo{author}{Verma, I.}, \bibinfo{author}{Dey, L.}, \bibinfo{author}{Bhardwaj, S.}, \bibinfo{year}{2024}.
\newblock \bibinfo{title}{Knowledge graph aided llm based esg question-answering from news} .
%Type = Article
\bibitem[{Gy{\"o}ny{\"o}rov{\'a} et~al.(2023)Gy{\"o}ny{\"o}rov{\'a}, Stacho{\v{n}} and Sta{\v{s}}ek}]{gyonyorova2023esg}
\bibinfo{author}{Gy{\"o}ny{\"o}rov{\'a}, L.}, \bibinfo{author}{Stacho{\v{n}}, M.}, \bibinfo{author}{Sta{\v{s}}ek, D.}, \bibinfo{year}{2023}.
\newblock \bibinfo{title}{Esg ratings: relevant information or misleading clue? evidence from the s\&p global 1200}.
\newblock \bibinfo{journal}{Journal of sustainable finance \& investment} \bibinfo{volume}{13}, \bibinfo{pages}{1075--1109}.
%Type = Article
\bibitem[{Hamilton et~al.(2024)Hamilton, Nayak, Bo{\v{z}}i{\'c} and Longo}]{hamilton2024neurosurvey}
\bibinfo{author}{Hamilton, K.}, \bibinfo{author}{Nayak, A.}, \bibinfo{author}{Bo{\v{z}}i{\'c}, B.}, \bibinfo{author}{Longo, L.}, \bibinfo{year}{2024}.
\newblock \bibinfo{title}{Is neuro-symbolic ai meeting its promises in natural language processing? a structured review}.
\newblock \bibinfo{journal}{Semantic Web} \bibinfo{volume}{15}, \bibinfo{pages}{1265--1306}.
%Type = Article
\bibitem[{Heichl and Hirsch(2023)}]{HEICHL2023textmining}
\bibinfo{author}{Heichl, V.}, \bibinfo{author}{Hirsch, S.}, \bibinfo{year}{2023}.
\newblock \bibinfo{title}{Sustainable fingerprint -- using textual analysis to detect how listed eu firms report about esg topics}.
\newblock \bibinfo{journal}{Journal of Cleaner Production} \bibinfo{volume}{426}, \bibinfo{pages}{138960}.
%Type = Article
\bibitem[{Henriksson et~al.(2019)Henriksson, Livnat, Pfeifer and Stumpp}]{henriksson2019integrating}
\bibinfo{author}{Henriksson, R.}, \bibinfo{author}{Livnat, J.}, \bibinfo{author}{Pfeifer, P.}, \bibinfo{author}{Stumpp, M.}, \bibinfo{year}{2019}.
\newblock \bibinfo{title}{Integrating esg in portfolio construction}.
\newblock \bibinfo{journal}{The Journal of Portfolio Management} \bibinfo{volume}{45}, \bibinfo{pages}{67--81}.
%Type = Inproceedings
\bibitem[{Huang and Chang(2023)}]{huang2022llmreasoning}
\bibinfo{author}{Huang, J.}, \bibinfo{author}{Chang, K.C.C.}, \bibinfo{year}{2023}.
\newblock \bibinfo{title}{Towards reasoning in large language models: A survey}, in: \bibinfo{booktitle}{Findings of the Association for Computational Linguistics: ACL 2023}, \bibinfo{publisher}{Association for Computational Linguistics}, \bibinfo{address}{Toronto, Canada}. pp. \bibinfo{pages}{1049--1065}.
%Type = Article
\bibitem[{Ignatov(2023)}]{ignatov2023esgdict}
\bibinfo{author}{Ignatov, K.}, \bibinfo{year}{2023}.
\newblock \bibinfo{title}{When esg talks: Esg tone of 10-k reports and its significance to stock markets}.
\newblock \bibinfo{journal}{International Review of Financial Analysis} \bibinfo{volume}{89}, \bibinfo{pages}{102745}.
%Type = Article
\bibitem[{Ji et~al.(2021)Ji, Pan, Cambria, Marttinen and Philip}]{ji2021KGsurvey}
\bibinfo{author}{Ji, S.}, \bibinfo{author}{Pan, S.}, \bibinfo{author}{Cambria, E.}, \bibinfo{author}{Marttinen, P.}, \bibinfo{author}{Philip, S.Y.}, \bibinfo{year}{2021}.
\newblock \bibinfo{title}{A survey on knowledge graphs: Representation, acquisition, and applications}.
\newblock \bibinfo{journal}{IEEE transactions on neural networks and learning systems} \bibinfo{volume}{33}, \bibinfo{pages}{494--514}.
%Type = Inproceedings
\bibitem[{Kang and El~Maarouf(2022)}]{kang2022finsim4terms}
\bibinfo{author}{Kang, J.}, \bibinfo{author}{El~Maarouf, I.}, \bibinfo{year}{2022}.
\newblock \bibinfo{title}{Finsim4-esg shared task: Learning semantic similarities for the financial domain. extended edition to esg insights}, in: \bibinfo{booktitle}{Proceedings of the Fourth Workshop on Financial Technology and Natural Language Processing (FinNLP)}, pp. \bibinfo{pages}{211--217}.
%Type = Article
\bibitem[{Lee and Seung(1999)}]{nmflee1999}
\bibinfo{author}{Lee, D.D.}, \bibinfo{author}{Seung, H.S.}, \bibinfo{year}{1999}.
\newblock \bibinfo{title}{Learning the parts of objects by non-negative matrix factorization}.
\newblock \bibinfo{journal}{nature} \bibinfo{volume}{401}, \bibinfo{pages}{788--791}.
%Type = Article
\bibitem[{Li and Leonas(2022)}]{li2022sustainabilityfashion}
\bibinfo{author}{Li, J.}, \bibinfo{author}{Leonas, K.K.}, \bibinfo{year}{2022}.
\newblock \bibinfo{title}{Sustainability topic trends in the textile and apparel industry: a text mining-based magazine article analysis}.
\newblock \bibinfo{journal}{Journal of Fashion Marketing and Management: An International Journal} \bibinfo{volume}{26}, \bibinfo{pages}{67--87}.
%Type = Article
\bibitem[{Liu and Singh(2004)}]{liu2004conceptnet}
\bibinfo{author}{Liu, H.}, \bibinfo{author}{Singh, P.}, \bibinfo{year}{2004}.
\newblock \bibinfo{title}{Conceptnet---a practical commonsense reasoning tool-kit}.
\newblock \bibinfo{journal}{BT technology journal} \bibinfo{volume}{22}, \bibinfo{pages}{211--226}.
%Type = Article
\bibitem[{Machado et~al.(2021)Machado, Dias and Fonseca}]{machado2021gri}
\bibinfo{author}{Machado, B.A.A.}, \bibinfo{author}{Dias, L.C.P.}, \bibinfo{author}{Fonseca, A.}, \bibinfo{year}{2021}.
\newblock \bibinfo{title}{Transparency of materiality analysis in gri-based sustainability reports}.
\newblock \bibinfo{journal}{Corporate Social Responsibility and Environmental Management} \bibinfo{volume}{28}, \bibinfo{pages}{570--580}.
%Type = Article
\bibitem[{Malesios et~al.(2021)Malesios, De, Moursellas, Dey and Evangelinos}]{malesios2021sustainabilityframework1}
\bibinfo{author}{Malesios, C.}, \bibinfo{author}{De, D.}, \bibinfo{author}{Moursellas, A.}, \bibinfo{author}{Dey, P.K.}, \bibinfo{author}{Evangelinos, K.}, \bibinfo{year}{2021}.
\newblock \bibinfo{title}{Sustainability performance analysis of small and medium sized enterprises: Criteria, methods and framework}.
\newblock \bibinfo{journal}{Socio-Economic Planning Sciences} \bibinfo{volume}{75}, \bibinfo{pages}{100993}.
%Type = Article
\bibitem[{Mandas et~al.(2023)Mandas, Lahmar, Piras and {De Lisa}}]{MANDAS2023esgtopicmodel}
\bibinfo{author}{Mandas, M.}, \bibinfo{author}{Lahmar, O.}, \bibinfo{author}{Piras, L.}, \bibinfo{author}{{De Lisa}, R.}, \bibinfo{year}{2023}.
\newblock \bibinfo{title}{Esg in the financial industry: What matters for rating analysts?}
\newblock \bibinfo{journal}{Research in International Business and Finance} \bibinfo{volume}{66}, \bibinfo{pages}{102045}.
%Type = Inproceedings
\bibitem[{Mao et~al.(2023)Mao, Li, He, Ge and Cambria}]{mao2023metaproonline}
\bibinfo{author}{Mao, R.}, \bibinfo{author}{Li, X.}, \bibinfo{author}{He, K.}, \bibinfo{author}{Ge, M.}, \bibinfo{author}{Cambria, E.}, \bibinfo{year}{2023}.
\newblock \bibinfo{title}{{MetaPro Online}: A computational metaphor processing online system}, in: \bibinfo{booktitle}{Proceedings of the 61st Annual Meeting of the Association for Computational Linguistics (Volume 3: System Demonstrations)}, \bibinfo{publisher}{Association for Computational Linguistics}, \bibinfo{address}{Toronto, Canada}. pp. \bibinfo{pages}{127--135}.
%Type = Article
\bibitem[{McInnes et~al.(2018)McInnes, Healy, Saul and Grossberger}]{mcinnes2018umap}
\bibinfo{author}{McInnes, L.}, \bibinfo{author}{Healy, J.}, \bibinfo{author}{Saul, N.}, \bibinfo{author}{Grossberger, L.}, \bibinfo{year}{2018}.
\newblock \bibinfo{title}{Umap: Uniform manifold approximation and projection}.
\newblock \bibinfo{journal}{The Journal of Open Source Software} \bibinfo{volume}{3}, \bibinfo{pages}{861}.
%Type = Article
\bibitem[{Meuer et~al.(2020)Meuer, Koelbel and Hoffmann}]{meuer2020naturecorporatesus}
\bibinfo{author}{Meuer, J.}, \bibinfo{author}{Koelbel, J.}, \bibinfo{author}{Hoffmann, V.H.}, \bibinfo{year}{2020}.
\newblock \bibinfo{title}{On the nature of corporate sustainability}.
\newblock \bibinfo{journal}{Organization \& Environment} \bibinfo{volume}{33}, \bibinfo{pages}{319--341}.
%Type = Article
\bibitem[{Moreno and Caminero(2022)}]{moreno2022climaterisk}
\bibinfo{author}{Moreno, A.I.}, \bibinfo{author}{Caminero, T.}, \bibinfo{year}{2022}.
\newblock \bibinfo{title}{Application of text mining to the analysis of climate-related disclosures}.
\newblock \bibinfo{journal}{International Review of Financial Analysis} \bibinfo{volume}{83}, \bibinfo{pages}{102307}.
%Type = Article
\bibitem[{Ng et~al.(2025)Ng, Nguyen, Huang, Tai, Leong, Leong, Yong, Ngui, Susanto, Cheng et~al.}]{ng2025sealion}
\bibinfo{author}{Ng, R.}, \bibinfo{author}{Nguyen, T.N.}, \bibinfo{author}{Huang, Y.}, \bibinfo{author}{Tai, N.C.}, \bibinfo{author}{Leong, W.Y.}, \bibinfo{author}{Leong, W.Q.}, \bibinfo{author}{Yong, X.}, \bibinfo{author}{Ngui, J.G.}, \bibinfo{author}{Susanto, Y.}, \bibinfo{author}{Cheng, N.}, et~al., \bibinfo{year}{2025}.
\newblock \bibinfo{title}{Sea-lion: Southeast asian languages in one network}.
\newblock \bibinfo{journal}{arXiv preprint arXiv:2504.05747} .
%Type = Article
\bibitem[{Nguyen and Huynh(2022)}]{nguyen2022textualcoporatebankruptcy}
\bibinfo{author}{Nguyen, B.H.}, \bibinfo{author}{Huynh, V.N.}, \bibinfo{year}{2022}.
\newblock \bibinfo{title}{Textual analysis and corporate bankruptcy: A financial dictionary-based sentiment approach}.
\newblock \bibinfo{journal}{Journal of the Operational Research Society} \bibinfo{volume}{73}, \bibinfo{pages}{102--121}.
%Type = Article
\bibitem[{Nguyen(2020)}]{nguyen2020keysource}
\bibinfo{author}{Nguyen, D.T.T.}, \bibinfo{year}{2020}.
\newblock \bibinfo{title}{An empirical study on the impact of sustainability reporting on firm value}.
\newblock \bibinfo{journal}{Journal of Competitiveness} \bibinfo{volume}{12}, \bibinfo{pages}{119--135}.
%Type = Article
\bibitem[{Ni et~al.(2023)Ni, Bingler, Colesanti-Senni, Kraus, Gostlow, Schimanski, Stammbach, Vaghefi, Wang, Webersinke et~al.}]{ni2023chatreport}
\bibinfo{author}{Ni, J.}, \bibinfo{author}{Bingler, J.}, \bibinfo{author}{Colesanti-Senni, C.}, \bibinfo{author}{Kraus, M.}, \bibinfo{author}{Gostlow, G.}, \bibinfo{author}{Schimanski, T.}, \bibinfo{author}{Stammbach, D.}, \bibinfo{author}{Vaghefi, S.A.}, \bibinfo{author}{Wang, Q.}, \bibinfo{author}{Webersinke, N.}, et~al., \bibinfo{year}{2023}.
\newblock \bibinfo{title}{Chatreport: Democratizing sustainability disclosure analysis through llm-based tools}.
\newblock \bibinfo{journal}{arXiv preprint arXiv:2307.15770} .
%Type = Article
\bibitem[{Ong et~al.(2025a)Ong, Mao, Satapathy, Filho, Cambria, Sulaeman and Mengaldo}]{ong2024xnlpsusanalysis}
\bibinfo{author}{Ong, K.}, \bibinfo{author}{Mao, R.}, \bibinfo{author}{Satapathy, R.}, \bibinfo{author}{Filho, R.S.}, \bibinfo{author}{Cambria, E.}, \bibinfo{author}{Sulaeman, J.}, \bibinfo{author}{Mengaldo, G.}, \bibinfo{year}{2025}a.
\newblock \bibinfo{title}{Explainable natural language processing for corporate sustainability analysis}.
\newblock \bibinfo{journal}{Information Fusion} \bibinfo{volume}{115}, \bibinfo{pages}{102726}.
%Type = Article
\bibitem[{Ong et~al.(2025b)Ong, Mao, Varshney, Cambria and Mengaldo}]{ong2025towardsrobustesganalysis}
\bibinfo{author}{Ong, K.}, \bibinfo{author}{Mao, R.}, \bibinfo{author}{Varshney, D.}, \bibinfo{author}{Cambria, E.}, \bibinfo{author}{Mengaldo, G.}, \bibinfo{year}{2025}b.
\newblock \bibinfo{title}{Towards robust esg analysis against greenwashing risks: Aspect-action analysis with cross-category generalization}.
\newblock \bibinfo{journal}{arXiv preprint arXiv:2502.15821} .
%Type = Article
\bibitem[{Papoutsi and Sodhi(2020)}]{papoutsi2020susdisclosuresdescriptive}
\bibinfo{author}{Papoutsi, A.}, \bibinfo{author}{Sodhi, M.S.}, \bibinfo{year}{2020}.
\newblock \bibinfo{title}{Does disclosure in sustainability reports indicate actual sustainability performance?}
\newblock \bibinfo{journal}{Journal of Cleaner Production} \bibinfo{volume}{260}, \bibinfo{pages}{121049}.
%Type = Article
\bibitem[{Park et~al.(2024)Park, Kim and Rim}]{park2024exploringviolations}
\bibinfo{author}{Park, K.}, \bibinfo{author}{Kim, H.}, \bibinfo{author}{Rim, H.}, \bibinfo{year}{2024}.
\newblock \bibinfo{title}{Exploring variations in corporations’ communication after a ca versus csr crisis: A semantic network analysis of sustainability reports}.
\newblock \bibinfo{journal}{International Journal of Business Communication} \bibinfo{volume}{61}, \bibinfo{pages}{240--262}.
%Type = Inproceedings
\bibitem[{Pasch and Ehnes(2022)}]{pasch2022sentimentclass}
\bibinfo{author}{Pasch, S.}, \bibinfo{author}{Ehnes, D.}, \bibinfo{year}{2022}.
\newblock \bibinfo{title}{Nlp for responsible finance: Fine-tuning transformer-based models for esg}, in: \bibinfo{booktitle}{2022 IEEE International Conference on Big Data (Big Data)}, \bibinfo{organization}{IEEE}. pp. \bibinfo{pages}{3532--3536}.
%Type = Article
\bibitem[{Peterson(2009)}]{peterson2009knn}
\bibinfo{author}{Peterson, L.E.}, \bibinfo{year}{2009}.
\newblock \bibinfo{title}{K-nearest neighbor}.
\newblock \bibinfo{journal}{Scholarpedia} \bibinfo{volume}{4}, \bibinfo{pages}{1883}.
%Type = Article
\bibitem[{Pikatza-Gorrotxategi et~al.(2024)Pikatza-Gorrotxategi, Borregan-Alvarado, Ruiz-de-la Torre-Acha and Alvarez-Meaza}]{Naiara2024news}
\bibinfo{author}{Pikatza-Gorrotxategi, N.}, \bibinfo{author}{Borregan-Alvarado, J.}, \bibinfo{author}{Ruiz-de-la Torre-Acha, A.}, \bibinfo{author}{Alvarez-Meaza, I.}, \bibinfo{year}{2024}.
\newblock \bibinfo{title}{News and esg investment criteria: What's behind it?}
\newblock \bibinfo{journal}{Social Network Analysis and Mining} \bibinfo{volume}{14}, \bibinfo{pages}{47}.
%Type = Article
\bibitem[{Rafailov et~al.(2024)Rafailov, Sharma, Mitchell, Manning, Ermon and Finn}]{rafailov2024direct}
\bibinfo{author}{Rafailov, R.}, \bibinfo{author}{Sharma, A.}, \bibinfo{author}{Mitchell, E.}, \bibinfo{author}{Manning, C.D.}, \bibinfo{author}{Ermon, S.}, \bibinfo{author}{Finn, C.}, \bibinfo{year}{2024}.
\newblock \bibinfo{title}{Direct preference optimization: Your language model is secretly a reward model}.
\newblock \bibinfo{journal}{Advances in Neural Information Processing Systems} \bibinfo{volume}{36}.
%Type = Inproceedings
\bibitem[{Reimers and Gurevych(2019)}]{reimers2019sentenceBERT}
\bibinfo{author}{Reimers, N.}, \bibinfo{author}{Gurevych, I.}, \bibinfo{year}{2019}.
\newblock \bibinfo{title}{Sentence-{BERT}: Sentence embeddings using {S}iamese {BERT}-networks}, in: \bibinfo{booktitle}{Proceedings of the 2019 Conference on Empirical Methods in Natural Language Processing and the 9th International Joint Conference on Natural Language Processing (EMNLP-IJCNLP)}, \bibinfo{publisher}{Association for Computational Linguistics}. pp. \bibinfo{pages}{3982--3992}.
%Type = Inproceedings
\bibitem[{Shahapure and Nicholas(2020)}]{shahapure2020silhouette}
\bibinfo{author}{Shahapure, K.R.}, \bibinfo{author}{Nicholas, C.}, \bibinfo{year}{2020}.
\newblock \bibinfo{title}{Cluster quality analysis using silhouette score}, in: \bibinfo{booktitle}{2020 IEEE 7th international conference on data science and advanced analytics (DSAA)}, \bibinfo{organization}{IEEE}. pp. \bibinfo{pages}{747--748}.
%Type = Article
\bibitem[{Silva(2021)}]{silva2021corporateprescriptive}
\bibinfo{author}{Silva, S.}, \bibinfo{year}{2021}.
\newblock \bibinfo{title}{Corporate contributions to the sustainable development goals: An empirical analysis informed by legitimacy theory}.
\newblock \bibinfo{journal}{Journal of Cleaner Production} \bibinfo{volume}{292}, \bibinfo{pages}{125962}.
%Type = Article
\bibitem[{Smeuninx et~al.(2020)Smeuninx, De~Clerck and Aerts}]{smeuninx2020susreportstedious}
\bibinfo{author}{Smeuninx, N.}, \bibinfo{author}{De~Clerck, B.}, \bibinfo{author}{Aerts, W.}, \bibinfo{year}{2020}.
\newblock \bibinfo{title}{Measuring the readability of sustainability reports: A corpus-based analysis through standard formulae and nlp}.
\newblock \bibinfo{journal}{International Journal of Business Communication} \bibinfo{volume}{57}, \bibinfo{pages}{52--85}.
%Type = Inproceedings
\bibitem[{Spillo et~al.(2022)Spillo, Musto, De~Gemmis, Lops and Semeraro}]{spillo2022neurorecosys}
\bibinfo{author}{Spillo, G.}, \bibinfo{author}{Musto, C.}, \bibinfo{author}{De~Gemmis, M.}, \bibinfo{author}{Lops, P.}, \bibinfo{author}{Semeraro, G.}, \bibinfo{year}{2022}.
\newblock \bibinfo{title}{Knowledge-aware recommendations based on neuro-symbolic graph embeddings and first-order logical rules}, in: \bibinfo{booktitle}{Proceedings of the 16th ACM Conference on Recommender Systems}, \bibinfo{publisher}{Association for Computing Machinery}, \bibinfo{address}{New York, NY, USA}. p. \bibinfo{pages}{616–621}.
%Type = Article
\bibitem[{Srivastava and Sutton(2017)}]{srivastava2017prodlda}
\bibinfo{author}{Srivastava, A.}, \bibinfo{author}{Sutton, C.}, \bibinfo{year}{2017}.
\newblock \bibinfo{title}{Autoencoding variational inference for topic models}.
\newblock \bibinfo{journal}{arXiv preprint arXiv:1703.01488} .
%Type = Article
\bibitem[{Tian et~al.(2023)Tian, Cheng, Xue, Han and Shan}]{tian2023naturesusdataset}
\bibinfo{author}{Tian, J.}, \bibinfo{author}{Cheng, Q.}, \bibinfo{author}{Xue, R.}, \bibinfo{author}{Han, Y.}, \bibinfo{author}{Shan, Y.}, \bibinfo{year}{2023}.
\newblock \bibinfo{title}{A dataset on corporate sustainability disclosure}.
\newblock \bibinfo{journal}{Scientific Data} \bibinfo{volume}{10}, \bibinfo{pages}{182}.
%Type = Article
\bibitem[{Wang and Leskovec(2021)}]{wang2021lpa}
\bibinfo{author}{Wang, H.}, \bibinfo{author}{Leskovec, J.}, \bibinfo{year}{2021}.
\newblock \bibinfo{title}{Combining graph convolutional neural networks and label propagation}.
\newblock \bibinfo{journal}{ACM Transactions on Information Systems (TOIS)} \bibinfo{volume}{40}, \bibinfo{pages}{1--27}.
%Type = Article
\bibitem[{Wang et~al.(2020)Wang, Yuen, Wong and Li}]{WANG2020sdg}
\bibinfo{author}{Wang, X.}, \bibinfo{author}{Yuen, K.F.}, \bibinfo{author}{Wong, Y.D.}, \bibinfo{author}{Li, K.X.}, \bibinfo{year}{2020}.
\newblock \bibinfo{title}{How can the maritime industry meet sustainable development goals? an analysis of sustainability reports from the social entrepreneurship perspective}.
\newblock \bibinfo{journal}{Transportation Research Part D: Transport and Environment} \bibinfo{volume}{78}, \bibinfo{pages}{102173}.
%Type = Article
\bibitem[{Wang et~al.(2024)Wang, Peng and Yang}]{wang2024decodingfirmvalue}
\bibinfo{author}{Wang, Y.}, \bibinfo{author}{Peng, Y.}, \bibinfo{author}{Yang, C.}, \bibinfo{year}{2024}.
\newblock \bibinfo{title}{Decoding esg report narratives: Unveiling sustainable supply chain insights and impacts through textual analysis}.
\newblock \bibinfo{journal}{Corporate Social Responsibility and Environmental Management} .
%Type = Article
\bibitem[{Wei et~al.(2022)Wei, Tay, Bommasani, Raffel, Zoph, Borgeaud, Yogatama, Bosma, Zhou, Metzler et~al.}]{wei2022emergent}
\bibinfo{author}{Wei, J.}, \bibinfo{author}{Tay, Y.}, \bibinfo{author}{Bommasani, R.}, \bibinfo{author}{Raffel, C.}, \bibinfo{author}{Zoph, B.}, \bibinfo{author}{Borgeaud, S.}, \bibinfo{author}{Yogatama, D.}, \bibinfo{author}{Bosma, M.}, \bibinfo{author}{Zhou, D.}, \bibinfo{author}{Metzler, D.}, et~al., \bibinfo{year}{2022}.
\newblock \bibinfo{title}{Emergent abilities of large language models}.
\newblock \bibinfo{journal}{arXiv preprint arXiv:2206.07682} .
%Type = Inproceedings
\bibitem[{Wu et~al.(2023)Wu, Dong, Nguyen and Luu}]{wu2023ecrtm}
\bibinfo{author}{Wu, X.}, \bibinfo{author}{Dong, X.}, \bibinfo{author}{Nguyen, T.T.}, \bibinfo{author}{Luu, A.T.}, \bibinfo{year}{2023}.
\newblock \bibinfo{title}{Effective neural topic modeling with embedding clustering regularization}, in: \bibinfo{editor}{Krause, A.}, \bibinfo{editor}{Brunskill, E.}, \bibinfo{editor}{Cho, K.}, \bibinfo{editor}{Engelhardt, B.}, \bibinfo{editor}{Sabato, S.}, \bibinfo{editor}{Scarlett, J.} (Eds.), \bibinfo{booktitle}{Proceedings of the 40th International Conference on Machine Learning}, \bibinfo{publisher}{PMLR}. pp. \bibinfo{pages}{37335--37357}.
%Type = Inproceedings
\bibitem[{Wu et~al.(2022)Wu, Luu and Dong}]{wu2022tsctm}
\bibinfo{author}{Wu, X.}, \bibinfo{author}{Luu, A.T.}, \bibinfo{author}{Dong, X.}, \bibinfo{year}{2022}.
\newblock \bibinfo{title}{Mitigating data sparsity for short text topic modeling by topic-semantic contrastive learning}, in: \bibinfo{booktitle}{Proceedings of the 2022 Conference on Empirical Methods in Natural Language Processing}, \bibinfo{publisher}{Association for Computational Linguistics}. pp. \bibinfo{pages}{2748--2760}.
%Type = Inproceedings
\bibitem[{Wu et~al.(2024)Wu, Pan and Luu}]{wu2023topmost}
\bibinfo{author}{Wu, X.}, \bibinfo{author}{Pan, F.}, \bibinfo{author}{Luu, A.T.}, \bibinfo{year}{2024}.
\newblock \bibinfo{title}{Towards the {T}op{M}ost: A topic modeling system toolkit}, in: \bibinfo{editor}{Cao, Y.}, \bibinfo{editor}{Feng, Y.}, \bibinfo{editor}{Xiong, D.} (Eds.), \bibinfo{booktitle}{Proceedings of the 62nd Annual Meeting of the Association for Computational Linguistics (Volume 3: System Demonstrations)}, \bibinfo{publisher}{Association for Computational Linguistics}, \bibinfo{address}{Bangkok, Thailand}. pp. \bibinfo{pages}{31--41}.
%Type = Article
\bibitem[{Yang et~al.(2024)Yang, Jin, Tang, Han, Feng, Jiang, Zhong, Yin and Hu}]{yang2024harnessing}
\bibinfo{author}{Yang, J.}, \bibinfo{author}{Jin, H.}, \bibinfo{author}{Tang, R.}, \bibinfo{author}{Han, X.}, \bibinfo{author}{Feng, Q.}, \bibinfo{author}{Jiang, H.}, \bibinfo{author}{Zhong, S.}, \bibinfo{author}{Yin, B.}, \bibinfo{author}{Hu, X.}, \bibinfo{year}{2024}.
\newblock \bibinfo{title}{Harnessing the power of llms in practice: A survey on chatgpt and beyond}.
\newblock \bibinfo{journal}{ACM Transactions on Knowledge Discovery from Data} \bibinfo{volume}{18}, \bibinfo{pages}{1--32}.
%Type = Article
\bibitem[{Ye et~al.(2022)Ye, Kumar, Sing, Song and Wang}]{ye2022comprehensive}
\bibinfo{author}{Ye, Z.}, \bibinfo{author}{Kumar, Y.J.}, \bibinfo{author}{Sing, G.O.}, \bibinfo{author}{Song, F.}, \bibinfo{author}{Wang, J.}, \bibinfo{year}{2022}.
\newblock \bibinfo{title}{A comprehensive survey of graph neural networks for knowledge graphs}.
\newblock \bibinfo{journal}{IEEE Access} \bibinfo{volume}{10}, \bibinfo{pages}{75729--75741}.
%Type = Article
\bibitem[{Yue et~al.(2023)Yue, Mao, Wang, Hu and Cambria}]{yue2023knowlenetsarcasm}
\bibinfo{author}{Yue, T.}, \bibinfo{author}{Mao, R.}, \bibinfo{author}{Wang, H.}, \bibinfo{author}{Hu, Z.}, \bibinfo{author}{Cambria, E.}, \bibinfo{year}{2023}.
\newblock \bibinfo{title}{Knowlenet: Knowledge fusion network for multimodal sarcasm detection}.
\newblock \bibinfo{journal}{Information Fusion} \bibinfo{volume}{100}, \bibinfo{pages}{101921}.
%Type = Inproceedings
\bibitem[{Zhang et~al.(2023)Zhang, Mao, He and Cambria}]{zhang2023neuro}
\bibinfo{author}{Zhang, X.}, \bibinfo{author}{Mao, R.}, \bibinfo{author}{He, K.}, \bibinfo{author}{Cambria, E.}, \bibinfo{year}{2023}.
\newblock \bibinfo{title}{Neuro-symbolic sentiment analysis with dynamic word sense disambiguation}, in: \bibinfo{booktitle}{Findings of the Association for Computational Linguistics: EMNLP 2023}, pp. \bibinfo{pages}{8772--8783}.
%Type = Article
\bibitem[{Zheng et~al.(2023)Zheng, Chiang, Sheng, Zhuang, Wu, Zhuang, Lin, Li, Li, Xing et~al.}]{zheng2023llmjudging}
\bibinfo{author}{Zheng, L.}, \bibinfo{author}{Chiang, W.L.}, \bibinfo{author}{Sheng, Y.}, \bibinfo{author}{Zhuang, S.}, \bibinfo{author}{Wu, Z.}, \bibinfo{author}{Zhuang, Y.}, \bibinfo{author}{Lin, Z.}, \bibinfo{author}{Li, Z.}, \bibinfo{author}{Li, D.}, \bibinfo{author}{Xing, E.}, et~al., \bibinfo{year}{2023}.
\newblock \bibinfo{title}{Judging llm-as-a-judge with mt-bench and chatbot arena}.
\newblock \bibinfo{journal}{Advances in Neural Information Processing Systems} \bibinfo{volume}{36}, \bibinfo{pages}{46595--46623}.
%Type = Article
\bibitem[{Zhong et~al.(2023)Zhong, Ding, Liu, Du, Jin and Tao}]{zhong2023knowledgeabsa}
\bibinfo{author}{Zhong, Q.}, \bibinfo{author}{Ding, L.}, \bibinfo{author}{Liu, J.}, \bibinfo{author}{Du, B.}, \bibinfo{author}{Jin, H.}, \bibinfo{author}{Tao, D.}, \bibinfo{year}{2023}.
\newblock \bibinfo{title}{Knowledge graph augmented network towards multiview representation learning for aspect-based sentiment analysis}.
\newblock \bibinfo{journal}{IEEE Transactions on knowledge and data engineering} \bibinfo{volume}{35}, \bibinfo{pages}{10098--10111}.
%Type = Article
\bibitem[{Zhou et~al.(2021)Zhou, Wang and Yuen}]{zhou2021shipping}
\bibinfo{author}{Zhou, Y.}, \bibinfo{author}{Wang, X.}, \bibinfo{author}{Yuen, K.F.}, \bibinfo{year}{2021}.
\newblock \bibinfo{title}{Sustainability disclosure for container shipping: A text-mining approach}.
\newblock \bibinfo{journal}{Transport Policy} \bibinfo{volume}{110}, \bibinfo{pages}{465--477}.
%Type = Article
\bibitem[{Zhu et~al.(2024)Zhu, Wang, Chen, Qiao, Ou, Yao, Deng, Chen and Zhang}]{zhu2024llmsforKB}
\bibinfo{author}{Zhu, Y.}, \bibinfo{author}{Wang, X.}, \bibinfo{author}{Chen, J.}, \bibinfo{author}{Qiao, S.}, \bibinfo{author}{Ou, Y.}, \bibinfo{author}{Yao, Y.}, \bibinfo{author}{Deng, S.}, \bibinfo{author}{Chen, H.}, \bibinfo{author}{Zhang, N.}, \bibinfo{year}{2024}.
\newblock \bibinfo{title}{Llms for knowledge graph construction and reasoning: Recent capabilities and future opportunities}.
\newblock \bibinfo{journal}{World Wide Web} \bibinfo{volume}{27}, \bibinfo{pages}{58}.

\end{thebibliography}


% %% For authoryear reference style

% \bibitem[Lamport(1994)]{lamport94}
%   \textit{\LaTeX: a document preparation system},
%   Addison Wesley, Massachusetts,

% \end{thebibliography}
\end{document}